\documentclass{article}

\usepackage{PRIMEarxiv}
\usepackage{calc}
\usepackage[utf8]{inputenc} 
\usepackage[T1]{fontenc}    
\usepackage[colorlinks=true, linkcolor=black, citecolor=black, urlcolor=black]{hyperref}     
\usepackage{url}            
\usepackage{booktabs}       
\usepackage{amsfonts}       
\usepackage{nicefrac}       
\usepackage{microtype}      
\usepackage{lipsum}
\usepackage{fancyhdr}       
\usepackage{graphicx}       
\usepackage{adjustbox}
\usepackage{fontawesome5}
\graphicspath{{media/}}     
\usepackage{graphicx}
\usepackage{tikz}
\usetikzlibrary{shapes.geometric, arrows, positioning}
\usepackage{listings}
\usepackage{amsmath}
\usepackage{xcolor}
\usepackage{float}
\usepackage{tcolorbox}
\usepackage{geometry}
\usepackage{multicol}
\usepackage{multirow}
\usepackage{algorithm2e}
\usepackage{enumitem}
\usepackage{colortbl}
\usepackage{array} 
\usepackage{caption}
\usepackage{subcaption}
\tikzstyle{startstop} = [rectangle, rounded corners, minimum width=3cm, minimum height=1cm, text centered, draw=black, fill=red!30]
\tikzstyle{process} = [rectangle, minimum width=2cm, minimum height=0.5cm, text centered, draw=black, fill=blue!10]
\tikzstyle{decision} = [diamond, minimum width=2cm, minimum height=0.5cm, text centered, draw=black, fill=green!10]
\tikzstyle{arrow} = [thick,->,>=stealth]
\tikzstyle{dashedarrow} = [dashed,->,>=stealth]
\tikzstyle{curvedarrow} = [thick,->,>=stealth, looseness=1.5, bend left]
\tikzstyle{info} = [rectangle, minimum width=3cm, minimum height=1cm, text centered, draw=black, fill=cyan!30]

\tikzstyle{block} = [rectangle, rounded corners, minimum width=3cm, minimum height=1cm, text centered, draw=black, fill=blue!10]
\tikzstyle{cloud} = [rectangle, minimum width=2.5cm, minimum height=1cm, text centered, draw=black, fill=red!10]
\tikzstyle{arrow} = [thick,->,>=stealth]
\tikzstyle{collection} = [draw, hexagon, text centered, minimum width=2cm, fill=green!10]
\tikzstyle{llm} = [rectangle, rounded corners, minimum width=2.5cm, minimum height=1cm, text centered, draw=black, fill=green!20]

\title{Agentic Retrieval-Augmented Generation: A Survey on Agentic RAG
}

\author{
  Aditi Singh \\
  Department of Computer Science \\
  Cleveland State University \\
  Cleveland, OH, USA \\
  \texttt{a.singh22@csuohio.edu} \\
  \And
  Abul Ehtesham \\
  Kent State University\\
  Kent, OH, USA \\
  \texttt{aehtesha@kent.edu } \\
  \And
  Saket Kumar \\
  Khoury College of Computer Science \\
  Northeastern University \\
  Boston, MA, USA \\
  \texttt{kumar.sak@northeastern.edu} \\
  \And
  Tala Talaei Khoei \\
  Khoury College of Computer Science \\
  Roux Institute at Northeastern University \\
  Portland, ME, USA \\
  \texttt{t.talaeikhoei@northeastern.edu} \\
  \And
  Athanasios V. Vasilakos \\
  Center for AI Research \\
  University of Agder (UiA) \\
  Jon Lilletuns vei 9, 4879 Grimstad, Norway \\
  \texttt{th.vasilakos@gmail.com} \\
}
\begin{document}
\maketitle

\begin{abstract}
Large Language Models (LLMs) have significantly advanced artificial intelligence by enabling human-like text generation and natural language understanding. However, their reliance on static training data limits their ability to respond to dynamic, real-time queries, resulting in outdated or inaccurate outputs. Retrieval-Augmented Generation (RAG) has emerged as a solution, enhancing LLMs by integrating real-time data retrieval to provide contextually relevant and up-to-date responses. Despite its promise, traditional RAG systems are constrained by static workflows and lack the adaptability required for multi-step reasoning and complex task management. Agentic Retrieval-Augmented Generation (Agentic RAG) transcends these limitations by embedding autonomous AI agents into the RAG pipeline. These agents leverage agentic design patterns reflection, planning, tool use, and multi-agent collaboration to dynamically manage retrieval strategies, iteratively refine contextual understanding, and adapt workflows through clearly defined operational structures ranging from sequential steps to adaptive collaboration. This integration enables Agentic RAG systems to deliver unparalleled flexibility, scalability, and context-awareness across diverse applications. This paper presents an analytical survey of Agentic RAG\footnote{\textit{GitHub link}: \protect\href{https://github.com/asinghcsu/AgenticRAG-Survey}{https://github.com/asinghcsu/AgenticRAG-Survey}} systems. It traces the evolution of RAG paradigms, introduces a principled taxonomy of Agentic RAG architectures based on agent cardinality, control structure, autonomy, and knowledge representation, and provides a comparative analysis of design trade-offs across existing frameworks. The survey further examines real-world applications in domains including healthcare, finance, education, and enterprise document processing, and distills practical lessons learned for system designers and practitioners. Finally, it identifies key open research challenges related to evaluation, coordination, memory management, efficiency, and governance, outlining directions for future research.


\end{abstract}

\keywords{Large Language Models (LLMs) \and Artificial Intelligence (AI) \and Natural Language Understanding \and Retrieval-Augmented Generation (RAG) \and Agentic RAG \and Autonomous AI Agents \and Reflection \and Planning \and Tool Use \and Multi-Agent Collaboration \and Agentic Patterns \and Contextual Understanding \and Dynamic Adaptability \and Scalability \and Real-Time Data Retrieval \and Taxonomy of Agentic RAG \and Healthcare Applications \and Finance Applications \and Educational Applications \and Ethical AI Decision-Making \and Performance Optimization \and Multi-Step Reasoning}

\section{Introduction}

Large Language Models (LLMs)~\cite{minaee2024largelanguagemodelssurvey,SinghLLM,zhao2024surveylargelanguagemodels}, such as GPT-5, PaLM, and LLaMA, have transformed artificial intelligence by enabling human-like text generation and natural language understanding. These models drive innovation in conversational agents~\cite{dam2024completesurveyllmbasedai}, content creation, real-time translation, and multimodal generation tasks including text-to-image and text-to-video synthesis~\cite{2023singhtextvideo,singh2024prompt}. However, a central challenge remains: how LLMs can reliably reason over rapidly changing external knowledge rather than relying solely on static pre-training.

Despite these advances, LLMs remain constrained by static pre-training data, often producing outdated responses, hallucinated content~\cite{Huang_2024}, and failing to adapt to evolving real-world information.

Retrieval-Augmented Generation (RAG)~\cite{lee2024agentg,zhao2024rag} addresses these limitations by integrating external knowledge into generation at inference time. By combining LLM capabilities with retrieval mechanisms~\cite{jiang2023activeretrievalaugmentedgeneration} over vector databases~\cite{han2023comprehensivesurveyvectordatabase}, APIs, or web sources, RAG improves factual grounding, relevance, and timeliness. However, traditional RAG pipelines are typically static and linear, limiting complex multi-step reasoning, deep contextual understanding, and iterative response refinement. These limitations motivate autonomous control mechanisms that dynamically decide when to retrieve, how to reformulate queries, and when sufficient evidence has been collected.

In parallel, advances in agentic AI~\cite{SAPKOTA2026103599} have introduced intelligent agents capable of perception, reasoning, and autonomous task execution~\cite{anthropic2024agents}. These agents leverage design patterns---reflection, planning, tool use, and multi-agent collaboration---to dynamically decompose tasks and adapt strategies. Such capabilities are directly relevant to RAG, as retrieval and reasoning decisions can themselves be treated as agentic tasks. These patterns are further structured through workflow patterns~\cite{anthropic2024agents,Bandara2025PracticalGuideAgenticAI}, including prompt chaining, routing, parallelization, orchestrator--worker models, and evaluator--optimizer loops, governing how multiple prompts, models, or agents are coordinated at the system level.

\begin{figure}
    \centering
    \includegraphics[width=\linewidth]{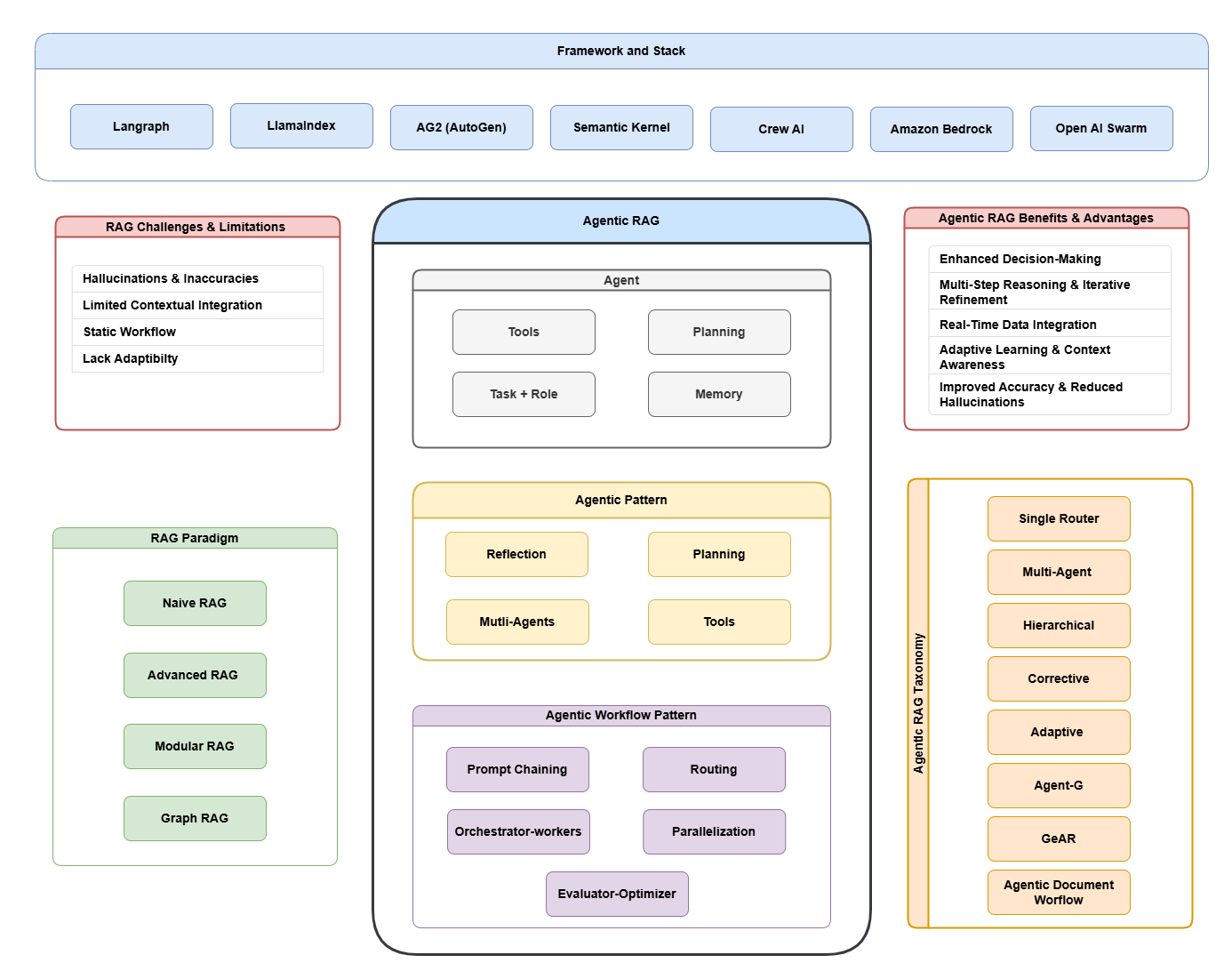}
    \caption{An Overview of Agentic RAG}
    \label{fig:oviewview_agentic_rag}
\end{figure}

The convergence of RAG and agentic intelligence has given rise to \emph{Agentic Retrieval-Augmented Generation (Agentic RAG)}~\cite{ravuru2024agenticrag}, which integrates autonomous agents directly into the RAG pipeline to enable dynamic retrieval, iterative context refinement, and adaptive workflow orchestration~\cite{huang2023reasoninglargelanguagemodels}. This introduces an explicit control layer that guides how the system reasons over external evidence---coordinating retrieval, validating information, invoking tools, and refining responses---making Agentic RAG effective for complex, multi-domain tasks requiring precision and adaptability, as illustrated in Fig.~\ref{fig:oviewview_agentic_rag}.

Despite rapidly growing interest, the Agentic RAG field remains fragmented, with diverse architectures, inconsistent terminology, and limited comparative understanding. Existing work focuses on individual frameworks or application-specific implementations, and no unified taxonomy currently organizes Agentic RAG systems along principled design dimensions.

This paper presents an analytical survey aimed at structuring this emerging field. It reviews the evolution of RAG paradigms, including naïve, modular, and graph-based RAG~\cite{peng2024graphretrievalaugmentedgenerationsurvey}, and their transition toward agentic systems. A principled taxonomy of Agentic RAG architectures is introduced based on agent cardinality, control structure, autonomy, and knowledge representation, followed by a comparative analysis of design trade-offs. The survey examines applications across healthcare~\cite{singh2024healthcare,gupta2024digitaldiagnosticspotentiallarge}, finance, and education~\cite{singheducation2025}, distills practical lessons for system designers, and identifies open research challenges to guide future work.

The structure of this paper is as follows: Section 2 introduces RAG and its evolution, highlighting the limitations of traditional approaches. Section 3 elaborates on the principles of agentic intelligence and agentic patterns. Section 4 elaborates on agentic workflow patterns. Section 5 provides a taxonomy of Agentic RAG systems, including single-agent, multi-agent, and graph-based frameworks. Section 6 provides a comparative analysis of Agentic RAG frameworks. Section 7 examines applications of Agentic RAG, while Section 8 discusses implementation tools and frameworks. Section 9 provides a comparative analysis of Agentic RAG frameworks. Section 10 discusses the lessons learned and practical guidance. Section 11 focuses on benchmarks and datasets, Section 12 discusses open research issues and future challenges, and Section 13 concludes with future directions for Agentic RAG systems.

\section{Foundations of Retrieval-Augmented Generation}

\subsection{Overview of Retrieval-Augmented Generation (RAG)}

Retrieval-Augmented Generation (RAG) represents a significant advancement in the field of artificial intelligence, combining the generative capabilities of Large Language Models (LLMs) with real-time data retrieval. While LLMs have demonstrated remarkable capabilities in natural language processing, their reliance on static pre-trained data often results in outdated or incomplete responses. RAG addresses this limitation by dynamically retrieving relevant information from external sources and incorporating it into the generative process, enabling contextually accurate and up-to-date outputs.

\subsection{Core Components of RAG}

\begin{figure}[htbp]
    \centering
    \includegraphics[width=0.8\linewidth]{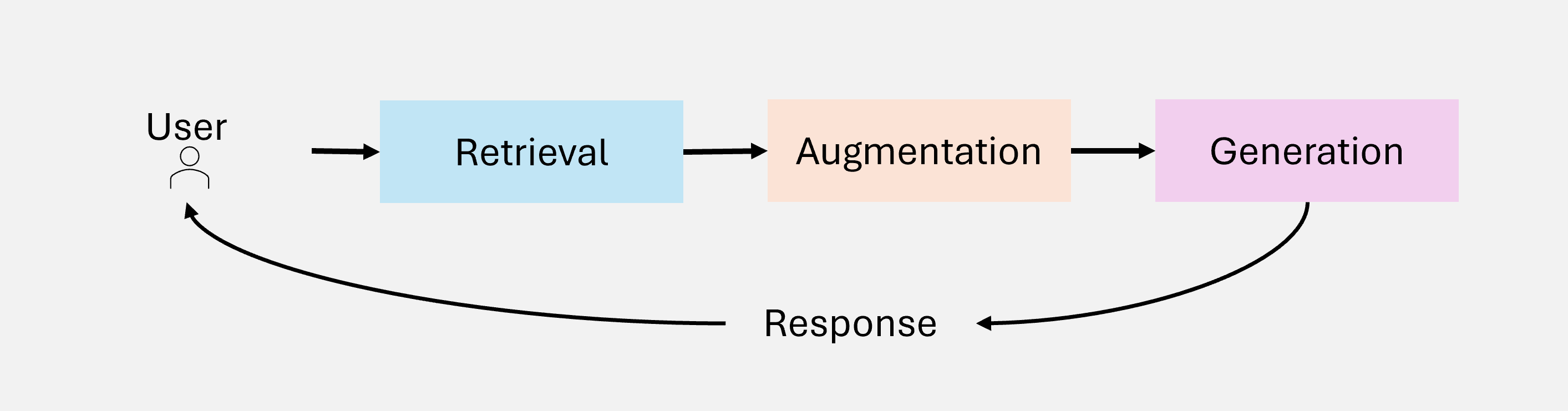}
    \caption{Core Components of RAG}
    \label{fig:rag_component}
\end{figure}

The architecture of RAG systems integrates three primary components (Figure\ref{fig:rag_component}):
\begin{itemize}
    \item \textbf{Retrieveal}: Responsible for querying external data sources such as knowledge bases, APIs, or vector databases. Advanced retrievers leverage dense vector search and transformer-based models to improve retrieval precision and semantic relevance.
    \item \textbf{Augmentation}: Processes retrieved data, extracting and summarizing the most relevant information to align with the query context.
    \item \textbf{Generation}: Combines retrieved information with the LLM’s pre-trained knowledge to generate coherent, contextually appropriate responses.
\end{itemize}

\subsection{Evolution of RAG Paradigms}

The field of Retrieval-Augmented Generation (RAG) has evolved significantly to address the increasing complexity of real-world applications, where contextual accuracy, scalability, and multi-step reasoning are critical. What began as simple keyword-based retrieval has transitioned into sophisticated, modular, and adaptive systems capable of integrating diverse data sources and autonomous decision-making processes. This evolution underscores the growing need for RAG systems to handle complex queries efficiently and effectively.

This section examines the progression of RAG paradigms, presenting key stages of development—Naïve RAG, Advanced RAG, Modular RAG, Graph RAG, and Agentic RAG alongside their defining characteristics, strengths, and limitations. By understanding the evolution of these paradigms, readers can appreciate the advancements made in retrieval and generative capabilities and their application in various domains

\subsubsection{Naïve RAG}

\begin{figure}[htbp]
    \centering
    \includegraphics[width=0.6\linewidth]{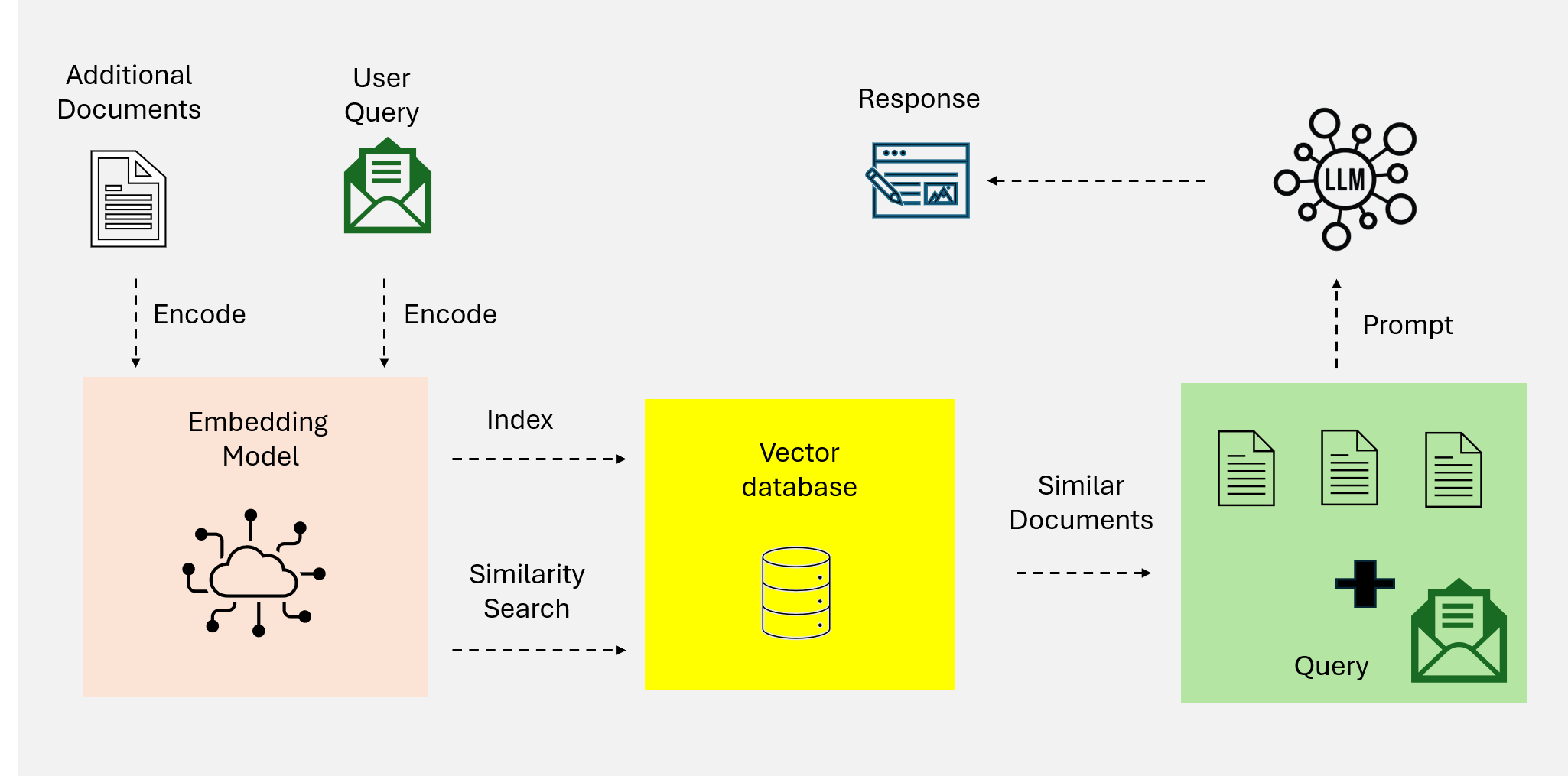}
    \caption{An Overview of Naive RAG.}
    \label{fig:naive_rag}
\end{figure}

Naïve RAG \cite{gao2024retrievalaugmentedgenerationlargelanguage} represents the foundational implementation of retrieval-augmented generation. Figure \ref{fig:naive_rag} illustrates the simple retrieve-read workflow of Naive RAG, focusing on keyword-based retrieval and static datasets.. These systems rely on simple keyword-based retrieval techniques, such as TF-IDF and BM25, to fetch documents from static datasets. The retrieved documents are then used to augment the language model’s generative capabilities.

Naïve RAG is characterized by its simplicity and ease of implementation, making it suitable for tasks involving fact-based queries with minimal contextual complexity. However, it suffers from several limitations:
\begin{itemize}
    \item \textbf{Lack of Contextual Awareness}: Retrieved documents often fail to capture the semantic nuances of the query due to reliance on lexical matching rather than semantic understanding.
    \item \textbf{Fragmented Outputs}: The absence of advanced preprocessing or contextual integration often leads to disjointed or overly generic responses.
    \item \textbf{Scalability Issues}: Keyword-based retrieval techniques struggle with large datasets, often failing to identify the most relevant information.
\end{itemize}

Despite these limitations, Naïve RAG systems provided a critical proof-of-concept for integrating retrieval with generation, laying the foundation for more sophisticated paradigms.

\subsubsection{Advanced RAG}

\begin{figure}[htbp]
    \centering
    \includegraphics[width=0.6\linewidth]{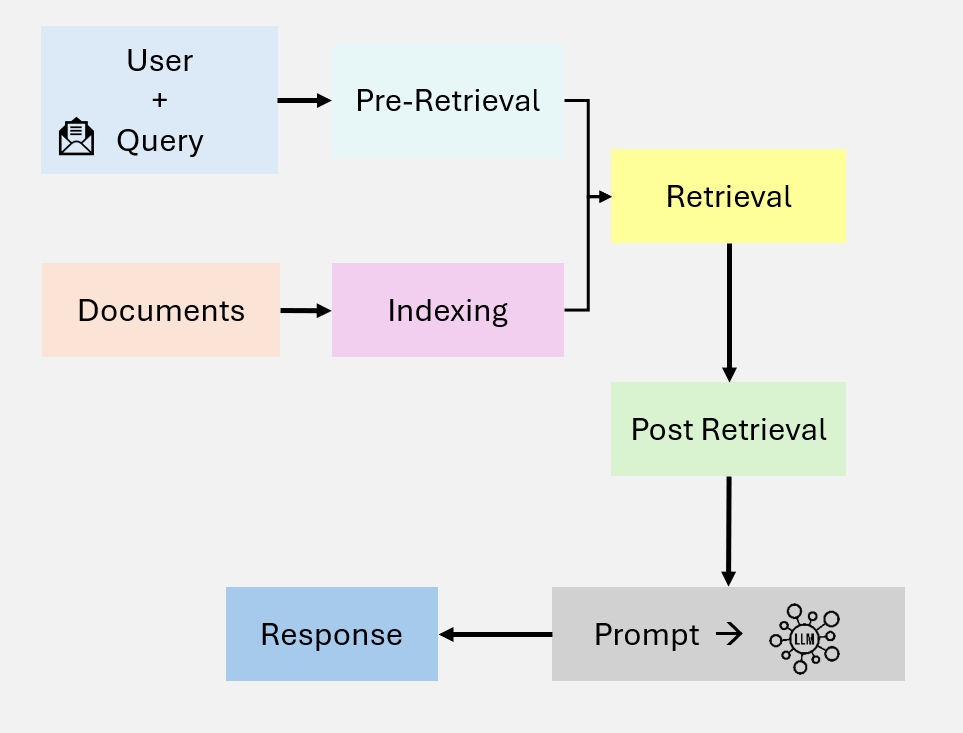}
    \caption{Overview of Advanced RAG}
    \label{fig:advanced_rag}
\end{figure}

Advanced RAG \cite{gao2024retrievalaugmentedgenerationlargelanguage} systems build upon the limitations of Naïve RAG by incorporating semantic understanding and enhanced retrieval techniques. Figure \ref{fig:advanced_rag} highlights the semantic enhancements in retrieval and the iterative, context-aware pipeline of Advanced RAG. These systems leverage dense retrieval models, such as Dense Passage Retrieval (DPR), and neural ranking algorithms to improve retrieval precision.

Key features of Advanced RAG include:
\begin{itemize}
    \item \textbf{Dense Vector Search}: Queries and documents are represented in high-dimensional vector spaces, enabling better semantic alignment between the user query and retrieved documents.
    \item \textbf{Contextual Re-Ranking}: Neural models re-rank retrieved documents to prioritize the most contextually relevant information.
    \item \textbf{Iterative Retrieval}: Advanced RAG introduces multi-hop retrieval mechanisms, enabling reasoning across multiple documents for complex queries.
\end{itemize}

These advancements make Advanced RAG suitable for applications requiring high precision and nuanced understanding, such as research synthesis and personalized recommendations. However, challenges such as computational overhead and limited scalability persist, particularly when dealing with large datasets or multi-step queries.

\subsubsection{Modular RAG}

\begin{figure}[H]
    \centering
    \includegraphics[width=0.6\linewidth]{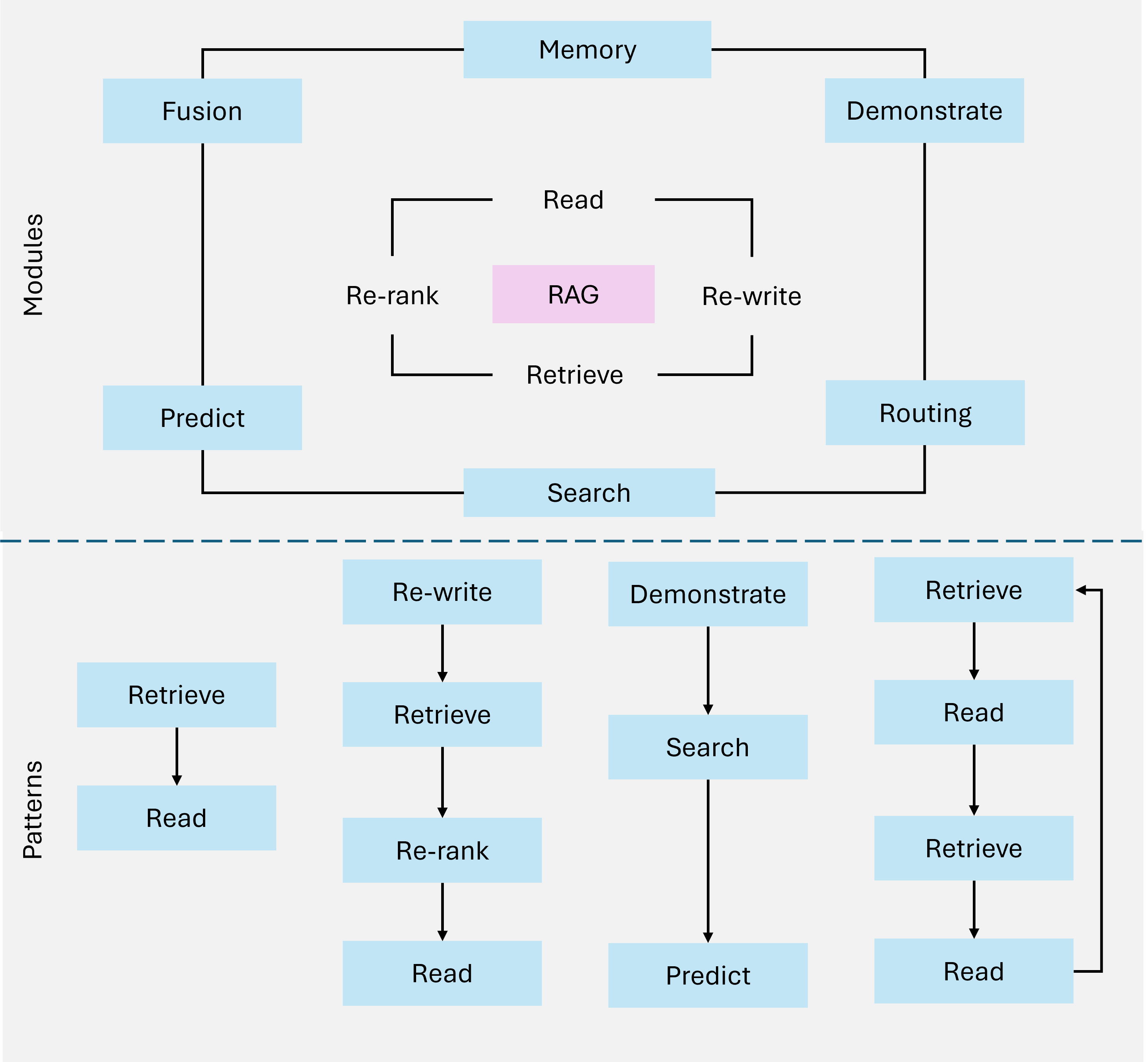}
    \caption{Overview of Modular RAG}
    \label{fig:modular_rag}
\end{figure}

Modular RAG \cite{gao2024retrievalaugmentedgenerationlargelanguage} represents the latest evolution in RAG paradigms, emphasizing flexibility and customization. These systems decompose the retrieval and generation pipeline into independent, reusable components, enabling domain-specific optimization and task adaptability. Figure \ref{fig:modular_rag} demonstrates the modular architecture, showcasing hybrid retrieval strategies, composable pipelines, and external tool integration.

Key innovations in Modular RAG include:
\begin{itemize}
    \item \textbf{Hybrid Retrieval Strategies}: Combining sparse retrieval methods (e.g., a sparse encoder-BM25) with dense retrieval techniques \cite{karpukhin2020densepassageretrievalopendomain} (e.g., DPR - Dense Passage Retrieval ) to maximize accuracy across diverse query types.
    \item \textbf{Tool Integration}: Incorporating external APIs, databases, or computational tools to handle specialized tasks, such as real-time data analysis or domain-specific computations.
    \item \textbf{Composable Pipelines}: Modular RAG enables retrievers, generators, and other components to be replaced, enhanced, or reconfigured independently, allowing high adaptability to specific use cases.
\end{itemize}

For instance, a Modular RAG system designed for financial analytics might retrieve live stock prices via APIs, analyze historical trends using dense retrieval, and generate actionable investment insights through a tailored language model. This modularity and customization make Modular RAG ideal for complex, multi-domain tasks, offering both scalability and precision.

\subsubsection{Graph RAG}

\begin{figure}[htbp]
    \centering
    \includegraphics[width=0.5\linewidth]{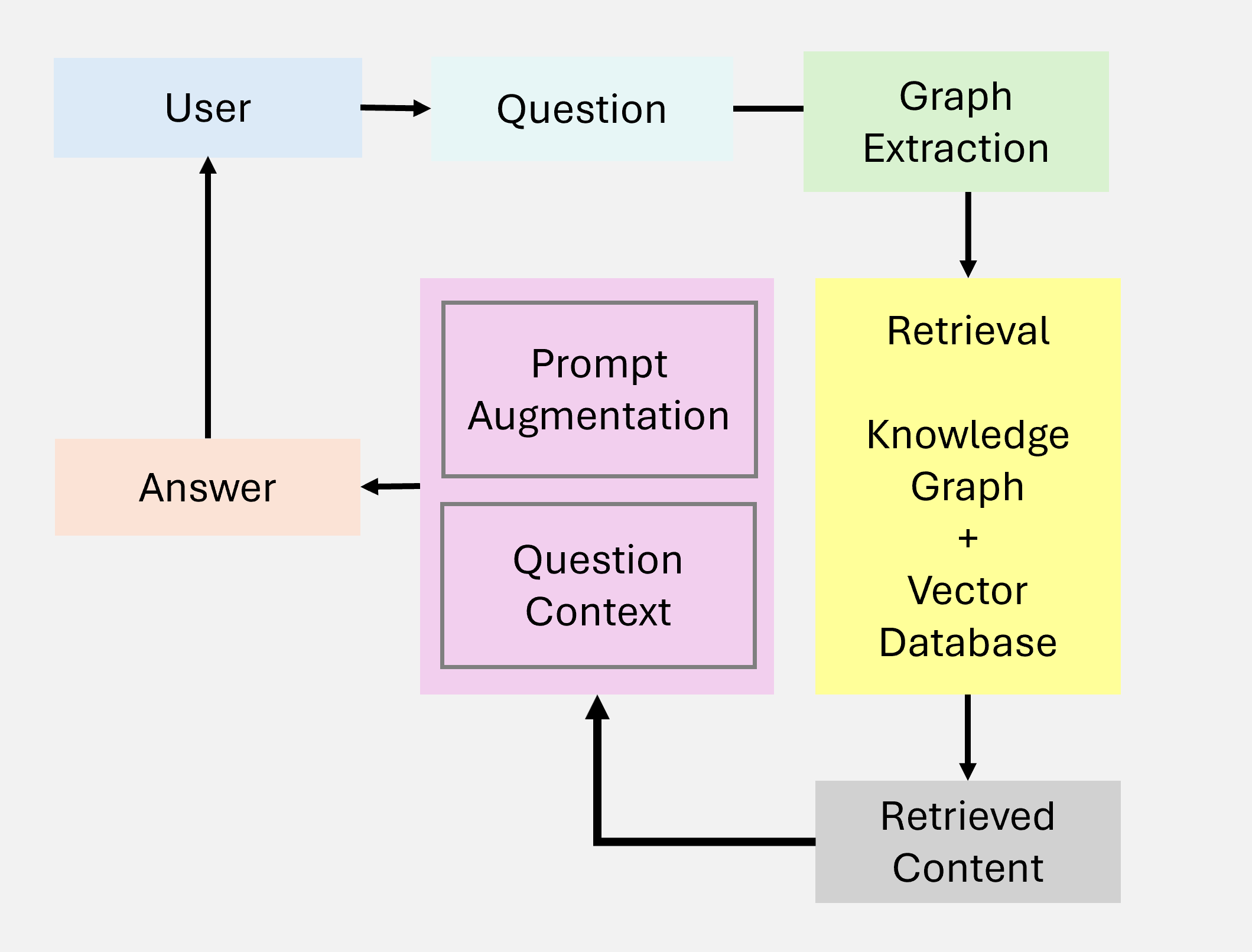}
    \caption{Overview of Graph RAG}
    \label{fig:graph_rag}
\end{figure}

Graph RAG \cite{peng2024graphretrievalaugmentedgenerationsurvey} extends traditional Retrieval-Augmented Generation systems by integrating graph-based data structures as illustrated in Figure \ref{fig:graph_rag}. These systems leverage the relationships and hierarchies within graph data to enhance multi-hop reasoning and contextual enrichment. By incorporating graph-based retrieval, Graph RAG enables richer and more accurate generative outputs, particularly for tasks requiring relational understanding.

Graph RAG is characterized by its ability to: \begin{itemize} \item \textbf{Node Connectivity}: Captures and reasons over relationships between entities. \item \textbf{Hierarchical Knowledge Management}: Handles structured and unstructured data through graph-based hierarchies. \item \textbf{Context Enrichment}: Adds relational understanding by leveraging graph-based pathways. \end{itemize}

However, Graph RAG has some limitations: \begin{itemize} \item \textbf{Limited Scalability}: The reliance on graph structures can restrict scalability, especially with extensive data sources. \item \textbf{Data Dependency}: High-quality graph data is essential for meaningful outputs, limiting its applicability in unstructured or poorly annotated datasets. \item \textbf{Complexity of Integration}: Integrating graph data with unstructured retrieval systems increases design and implementation complexity. \end{itemize}

Graph RAG is well-suited for applications such as healthcare diagnostics, legal research, and other domains where reasoning over structured relationships is crucial.

\subsubsection{Agentic RAG}

Agentic RAG represents a paradigm shift by introducing autonomous agents capable of dynamic decision-making and workflow optimization. Unlike static systems, Agentic RAG employs iterative refinement and adaptive retrieval strategies to address complex, real-time, and multi-domain queries. This paradigm leverages the modularity of retrieval and generation processes while introducing agent-based autonomy.

Key characteristics of Agentic RAG include: \begin{itemize} \item \textbf{Autonomous Decision-Making}: Agents independently evaluate and manage retrieval strategies based on query complexity. \item \textbf{Iterative Refinement}: Incorporates feedback loops to improve retrieval accuracy and response relevance. \item \textbf{Workflow Optimization}: Dynamically orchestrates tasks, enabling efficiency in real-time applications. \end{itemize}

Despite its advancements, Agentic RAG faces some challenges: \begin{itemize} \item \textbf{Coordination Complexity}: Managing interactions between agents requires sophisticated orchestration mechanisms. \item \textbf{Computational Overhead}: The use of multiple agents increases resource requirements for complex workflows. \item \textbf{Scalability Limitations}: While scalable, the dynamic nature of the system can strain computational resources for high query volumes. \end{itemize}

Agentic RAG excels in domains like customer support, financial analytics, and adaptive learning platforms, where dynamic adaptability and contextual precision are paramount.


\begin{table}[h]
\centering
\caption{Comparative Analysis of RAG Paradigms}
\renewcommand{\arraystretch}{1.4}
\begin{tabular}{|>{\centering\arraybackslash}m{3.5cm}|>{\centering\arraybackslash}m{5.5cm}|>{\centering\arraybackslash}m{6cm}|}
\hline
\rowcolor{lightgray} \textbf{Paradigm} & \textbf{Key Features} & \textbf{Strengths} \\ \hline
\textbf{Naïve RAG} & 
\begin{itemize}
    \item Keyword-based retrieval (e.g., TF-IDF, BM25)
\end{itemize} & 
\begin{itemize}
    \item Simple and easy to implement
    \item Suitable for fact-based queries
\end{itemize} \\ \hline

\textbf{Advanced RAG} & 
\begin{itemize}
    \item Dense retrieval models (e.g., DPR)
    \item Neural ranking and re-ranking
    \item Multi-hop retrieval
\end{itemize} & 
\begin{itemize}
    \item High precision retrieval
    \item Improved contextual relevance
\end{itemize} \\ \hline

\textbf{Modular RAG} & 
\begin{itemize}
    \item Hybrid retrieval (sparse and dense)
    \item Tool and API integration
    \item Composable, domain-specific pipelines
\end{itemize} & 
\begin{itemize}
    \item High flexibility and customization
    \item Suitable for diverse applications
    \item Scalable
\end{itemize} \\ \hline

\textbf{Graph RAG} & 
\begin{itemize}
    \item Integration of graph-based structures
    \item Multi-hop reasoning
    \item Contextual enrichment via nodes
\end{itemize} & 
\begin{itemize}
    \item Relational reasoning capabilities
    \item Mitigates hallucinations
    \item Ideal for structured data tasks
\end{itemize} \\ \hline

\textbf{Agentic RAG} & 
\begin{itemize}
    \item Autonomous agents
    \item Dynamic decision-making
    \item Iterative refinement and workflow optimization
\end{itemize} & 
\begin{itemize}
    \item Adaptable to real-time changes
    \item Scalable for multi-domain tasks
    \item High accuracy
\end{itemize} \\ \hline
\end{tabular}
\label{tab:rag_comparison}
\end{table}




\subsection{Challenges and Limitations of Traditional RAG Systems}

Traditional Retrieval-Augmented Generation (RAG) systems have significantly expanded the capabilities of Large Language Models (LLMs) by integrating real-time data retrieval. However, these systems still face critical challenges that hinder their effectiveness in complex, real-world applications. The most notable limitations revolve around \textbf{contextual integration}, \textbf{multi-step reasoning}, and \textbf{scalability and latency issues}.

\subsubsection{Contextual Integration}

Even when RAG systems successfully retrieve relevant information, they often struggle to seamlessly incorporate it into generated responses. The static nature of retrieval pipelines and limited contextual awareness lead to fragmented, inconsistent, or overly generic outputs.

\textbf{Example:}
 A query such as, \textit{"What are the latest advancements in Alzheimer's research and their implications for early-stage treatment?"} might yield relevant research papers and medical guidelines. However, traditional RAG systems often fail to synthesize these findings into a coherent explanation that connects the new treatments to specific patient scenarios.
Similarly, for a query like, \textit{"What are the best sustainable practices for small-scale agriculture in arid regions?"}, traditional systems might retrieve documents on general agricultural methods but overlook critical sustainability practices tailored to arid environments.

\subsubsection{Multi-Step Reasoning}

Many real-world queries require iterative or multi-hop reasoning—retrieving and synthesizing information across multiple steps. Traditional RAG systems are often ill-equipped to refine retrieval based on intermediate insights or user feedback, resulting in incomplete or disjointed responses.

\textbf{Example:} A complex query like, \textit{"What lessons from renewable energy policies in Europe can be applied to developing nations, and what are the potential economic impacts?"} demands the orchestration of multiple types of information, including policy data, contextualization for developing regions, and economic analysis. Traditional RAG systems typically fail to connect these disparate elements into a cohesive response.

\subsubsection{Scalability and Latency Issues}

As the volume of external data sources grows, querying and ranking large datasets becomes increasingly computationally intensive. This results in significant latency, which undermines the system's ability to provide timely responses in real-time applications.

\textbf{Example:}
In time-sensitive settings such as \textit{financial analytics} or \textit{live customer support}, delays caused by querying multiple databases or processing large document sets can hinder the system's overall utility. For example, a delay in retrieving market trends during high-frequency trading could result in missed opportunities.

\subsection{Agentic RAG: A Paradigm Shift}
Traditional RAG systems, with their static workflows and limited adaptability, often struggle to handle dynamic, 
multi-step reasoning and complex real-world tasks. These limitations have spurred the integration of agentic 
intelligence, resulting in Agentic RAG. By incorporating autonomous agents capable of 
dynamic decision-making, iterative reasoning, and adaptive retrieval strategies, Agentic RAG builds on the 
modularity of earlier paradigms while overcoming their inherent constraints. This evolution enables more complex, 
multi-domain tasks to be addressed with enhanced precision and contextual understanding, positioning Agentic RAG 
as a cornerstone for next-generation AI applications. In particular, Agentic RAG systems reduce latency through 
optimized workflows and refine outputs iteratively, tackling the very challenges that have historically hindered 
traditional RAG’s scalability and effectiveness.

\section{Core Principles and Background of Agentic Intelligence}

Agentic Intelligence forms the foundation of Agentic Retrieval-Augmented Generation (RAG) systems, enabling them to transcend the static and reactive nature of traditional RAG. By integrating autonomous agents capable of dynamic decision-making, iterative reasoning, and collaborative workflows, Agentic RAG systems exhibit enhanced adaptability and precision. This section explores the core principles underpinning agentic intelligence.

\paragraph{ Components of an AI Agent.}
In essence, an AI agent comprises (Figure. \ref{fig:ai_agents}):
\begin{itemize}
    \item \textbf{LLM (with defined Role and Task):} 
    Serves as the agent’s primary reasoning engine and dialogue interface. It interprets user queries, generates responses, and maintains coherence.
    \item \textbf{Memory (Short-Term and Long-Term):} 
    Captures context and relevant data across interactions. Short-term memory \cite{zhang2024surveymemorymechanismlarge} tracks immediate conversation state, while long-term memory \cite{zhang2024surveymemorymechanismlarge}stores accumulated knowledge and agent experiences.
    \item \textbf{Planning (Reflection \& Self-Critique):} 
    Guides the agent’s iterative reasoning process through reflection, query routing, or self-critique\cite{gou2024criticlargelanguagemodels}, ensuring that complex tasks are broken down effectively \cite{huang2024planningllmagents}.
    \item \textbf{Tools Vector Search, Web Search, APIs, etc.):} 
    Expands the agent’s capabilities beyond text generation, enabling access to external resources, real-time data, or specialized computations.
\end{itemize}

\begin{center}
     \centering
     \includegraphics[width=0.9\linewidth]{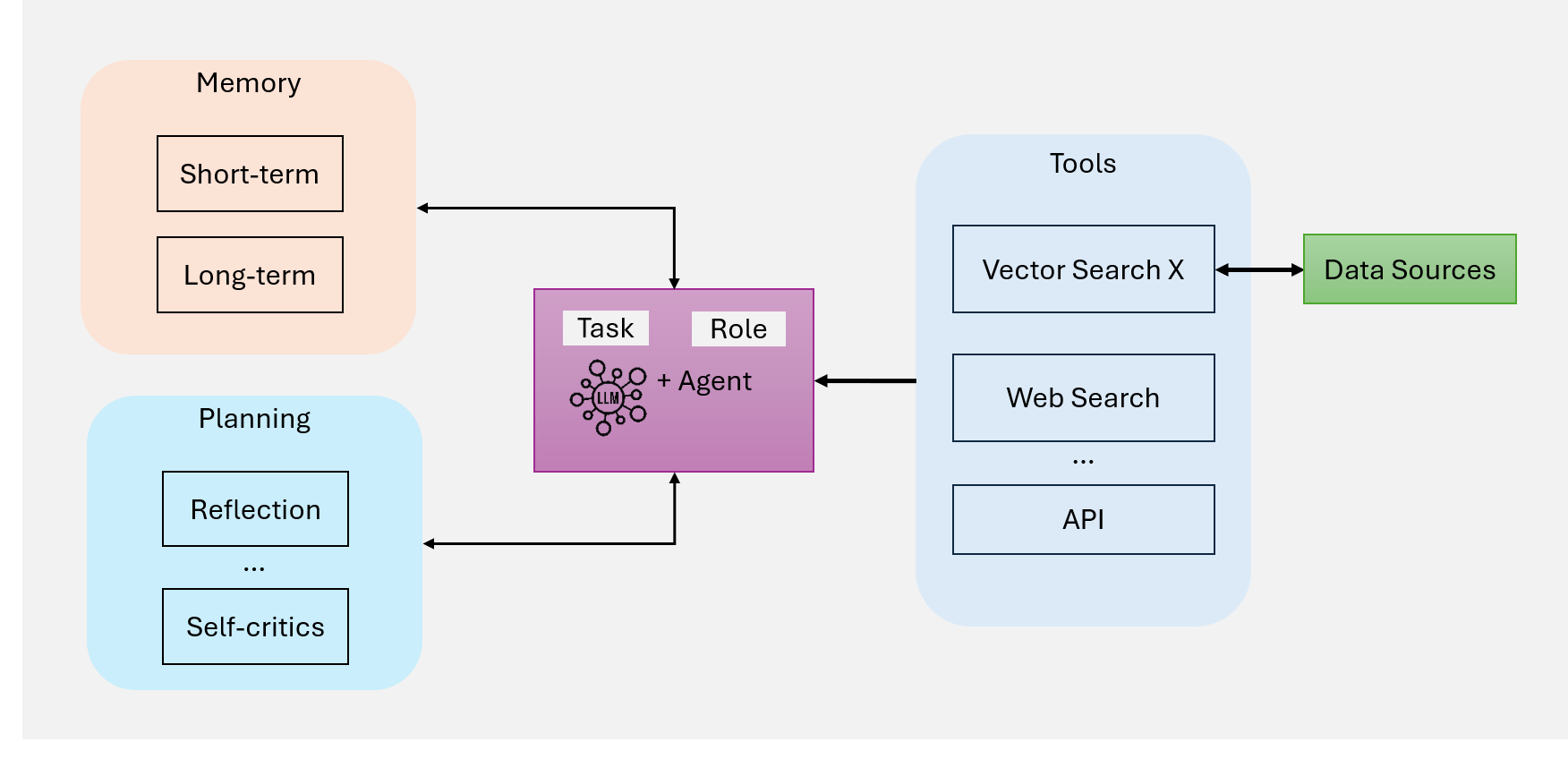}
    \captionof{figure}{An Overview of AI Agents}
    \label{fig:ai_agents}
\end{center}

Agentic Patterns \cite{SinghAgentic, 2025agenticpattern} provide structured methodologies that guide the behavior of agents in Agentic Retrieval-Augmented Generation (RAG) systems. These patterns enable agents to dynamically adapt, plan, and collaborate, ensuring that the system can handle complex, real-world tasks with precision and scalability. Four key patterns underpin agentic workflows:

\subsection{Reflection} 
\begin{figure}[H]
    \centering
    \includegraphics[width=0.4\linewidth]{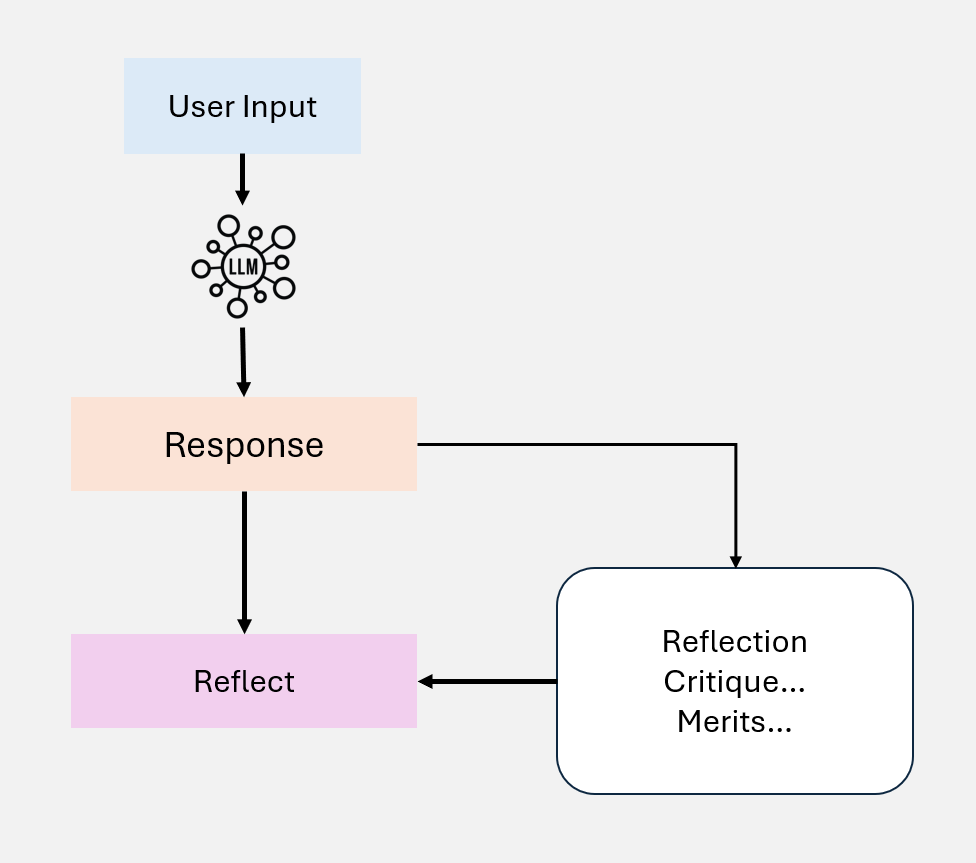}
    \caption{An Overview of Agentic Self- Reflection}
    \label{fig:agentic_self_reflection}
\end{figure}

Reflection is a foundational design pattern in agentic workflows, enabling agents to iteratively evaluate and refine their outputs. By incorporating self-feedback mechanisms, agents can identify and address errors, inconsistencies, and areas for improvement, enhancing performance across tasks like code generation, text production, and question answering ( as shown in Figure \ref{fig:agentic_self_reflection}). In practical use, Reflection involves prompting an agent to critique its outputs for correctness, style, and efficiency, then incorporating this feedback into subsequent iterations. External tools, such as unit tests or web searches, can further enhance this process by validating results and highlighting gaps.

In multi-agent systems, Reflection can involve distinct roles, such as one agent generating outputs while another critiques them, fostering collaborative improvement. For instance, in legal research, agents can iteratively refine responses by re-evaluating retrieved case law, ensuring accuracy and comprehensiveness. Reflection has demonstrated significant performance improvements in studies like \textit{Self-Refine} \cite{madaan2023selfrefineiterativerefinementselffeedback}, \textit{Reflexion} \cite{shinn2023reflexionlanguageagentsverbal}, and \textit{CRITIC} \cite{gou2024criticlargelanguagemodels}.

\subsection{Planning}
\begin{figure}[htbp]
    \centering
    \begin{subfigure}[b]{0.47\linewidth}
        \includegraphics[width=\linewidth]{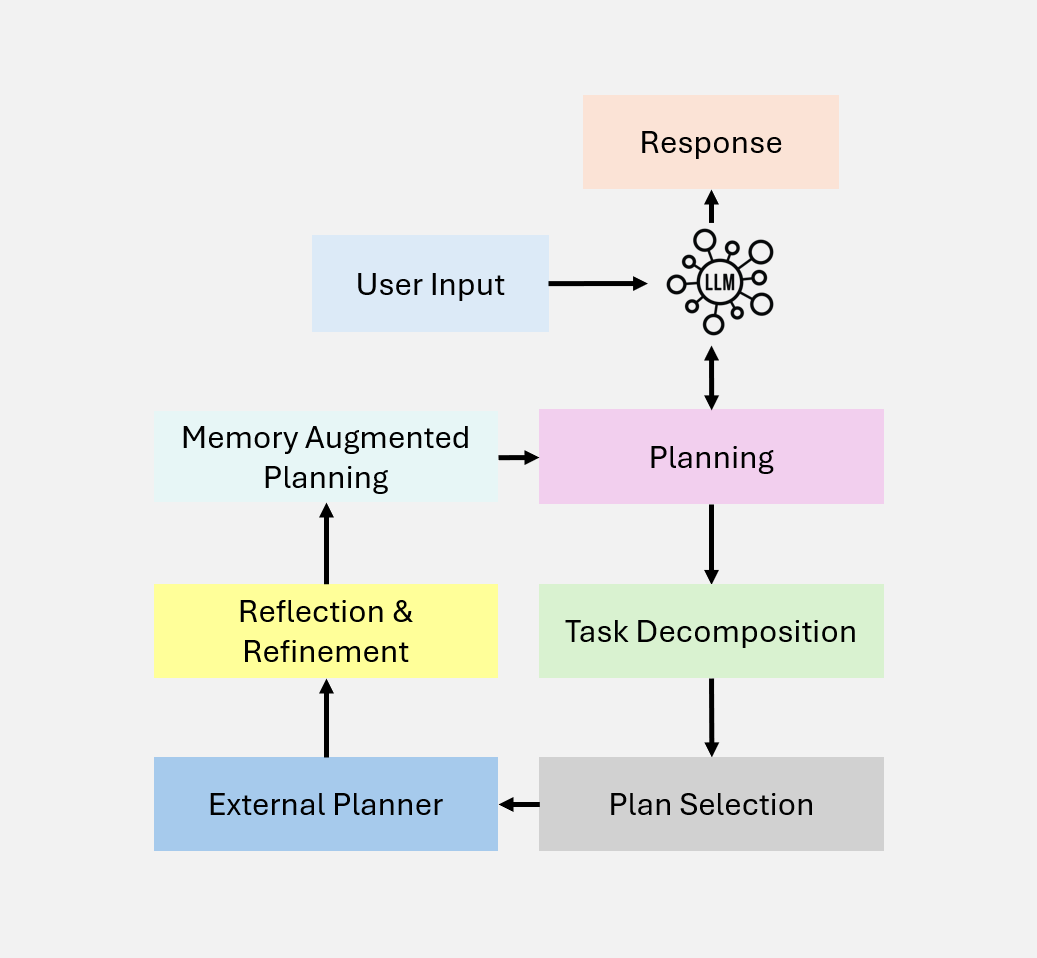}
        \caption{An Overview of Agentic Planning}
        \label{fig:agentic_planning}
    \end{subfigure}
    \hfill
    \begin{subfigure}[b]{0.47\linewidth}
        \includegraphics[width=\linewidth]{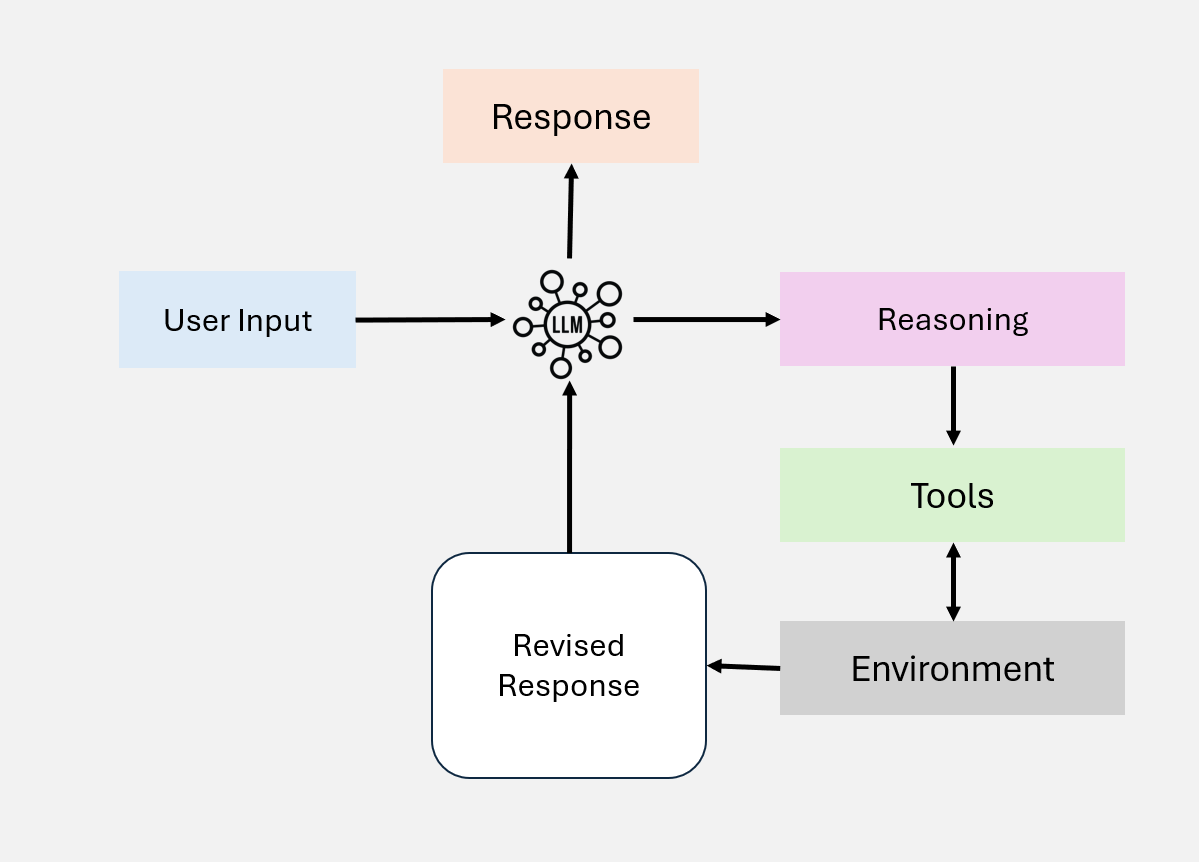}
        \caption{An Overview of Tool Use}
        \label{fig:agentic_tool_use}
    \end{subfigure}
    \caption{Overview of Agentic Planning and Tool Use}
    \label{fig:agentic_comparison}
\end{figure}

Planning \cite{huang2024planningllmagents} is a key design pattern in agentic workflows that enables agents to autonomously decompose complex tasks into smaller, manageable subtasks. This capability is essential for multi-hop reasoning and iterative problem-solving in dynamic and uncertain scenarios as shown in Figure \ref{fig:agentic_planning}. 

By leveraging planning, agents can dynamically determine the sequence of steps needed to accomplish a larger objective. This adaptability allows agents to handle tasks that cannot be predefined, ensuring flexibility in decision-making. While powerful, Planning can produce less predictable outcomes compared to deterministic workflows like Reflection. Planning is particularly suited for tasks that require dynamic adaptation, where predefined workflows are insufficient. As the technology matures, its potential to drive innovative applications across domains will continue to grow.


\subsection{Tool Use} 
Tool Use enables agents to extend their capabilities by interacting with external tools, APIs, or computational resources as illustrated in \ref{fig:agentic_tool_use}. This pattern allows agents to gather information, perform computations, and manipulate data beyond their pre-trained knowledge. By dynamically integrating tools into workflows, agents can adapt to complex tasks and provide more accurate and contextually relevant outputs.


Modern agentic workflows incorporate tool use for a variety of applications, including information retrieval, computational reasoning, and interfacing with external systems. The implementation of this pattern has evolved significantly with advancements like GPT-4’s function calling capabilities and systems capable of managing access to numerous tools. These developments facilitate sophisticated workflows where agents autonomously select and execute the most relevant tools for a given task.

While tool use significantly enhances agentic workflows, challenges remain in optimizing the selection of tools, particularly in contexts with a large number of available options. Techniques inspired by retrieval-augmented generation (RAG), such as heuristic-based selection, have been proposed to address this issue.






\subsection{Multi-Agent}
\begin{figure}[h]
    \centering
    \includegraphics[width=0.5\linewidth]{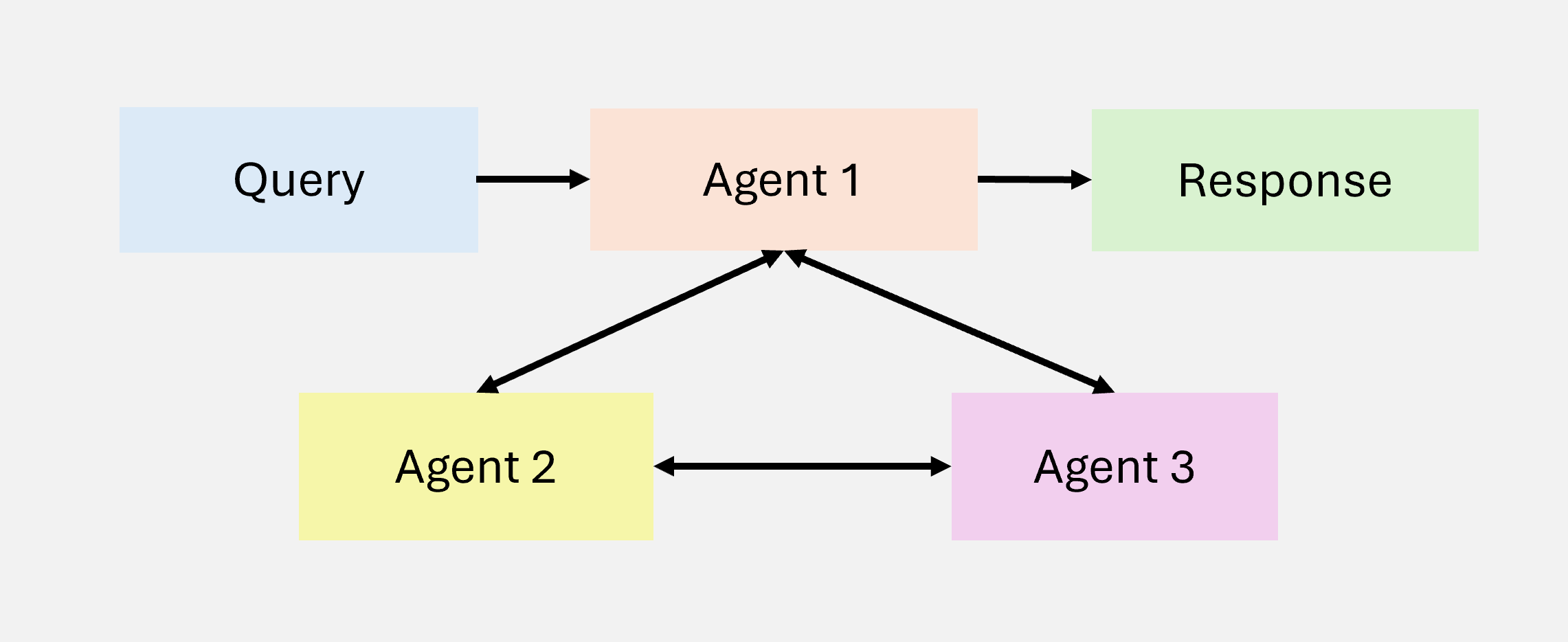}
    \caption{An Overview of MultiAgent }
    \label{fig:multi_agent_pattern}
\end{figure}

Multi-agent collaboration \cite{guo2024multiagent} is a key design pattern in agentic workflows that enables task specialization and parallel processing. Agents communicate and share intermediate results, ensuring the overall workflow remains efficient and coherent. By distributing subtasks among specialized agents, this pattern improves the scalability and adaptability of complex workflows. Multi-agent systems allow developers to decompose intricate tasks into smaller, manageable subtasks assigned to different agents. This approach not only enhances task performance but also provides a robust framework for managing complex interactions. Each agent operates with its own memory and workflow, which can include the use of tools, reflection, or planning, enabling dynamic and collaborative problem-solving (see Figure \ref{fig:multi_agent_pattern}).

While multi-agent collaboration offers significant potential, it is a less predictable design pattern compared to more mature workflows like Reflection and Tool Use. Nevertheless, emerging frameworks such as AutoGen, Crew AI, and LangGraph are providing new avenues for implementing effective multi-agent solutions.

These design patterns form the foundation for the success of Agentic RAG systems. By structuring workflows—from simple, sequential steps to more adaptive, collaborative processes—these patterns enable systems to dynamically adapt their retrieval and generative strategies to the diverse and ever-changing demands of real-world environments. Leveraging these patterns, agents are capable of handling iterative, context-aware tasks that significantly exceed the capabilities of traditional RAG systems.

\section{Agentic Workflow Patterns: Adaptive Strategies for Dynamic Collaboration}

Agentic workflow patterns, \cite{anthropic2024agents, langgraph2025workflows} structure LLM-based applications to optimize performance, accuracy, and efficiency. Different approaches are suitable depending on task complexity and processing requirements.

\subsection{Prompt Chaining: Enhancing Accuracy Through Sequential Processing}

Prompt chaining \cite{anthropic2024agents, langgraph2025workflows} decomposes a complex task into multiple steps, where each step builds upon the previous one. This structured approach improves accuracy by simplifying each subtask before moving forward. However, it may increase latency due to sequential processing.

\begin{figure}[h]
    \centering
    \includegraphics[width=\linewidth]{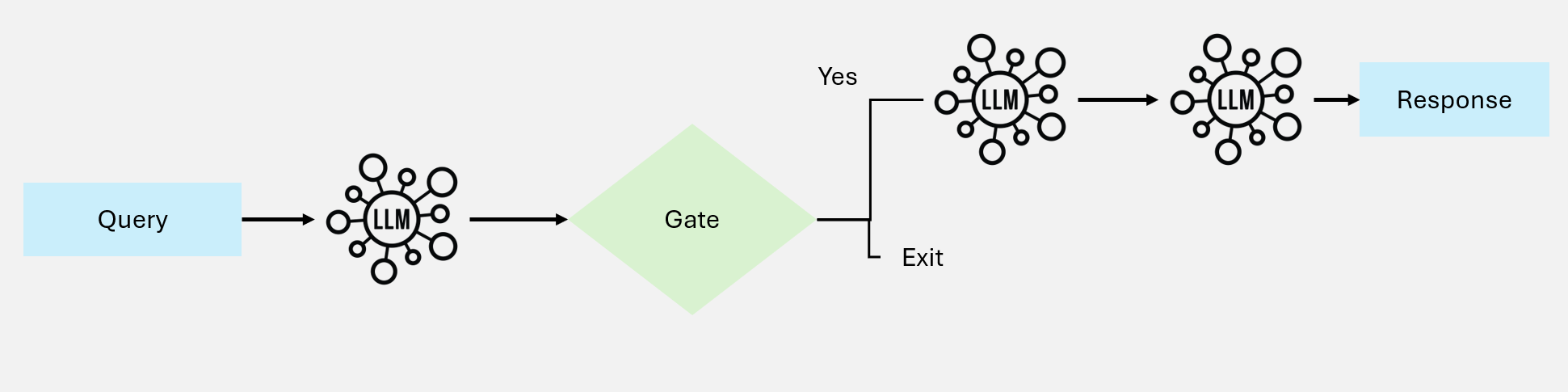}
    \caption{Illustration of Prompt Chaining Workflow}
    \label{fig:prompt_chaining}
\end{figure}

    \textbf{When to Use:} This workflow is most effective when a task can be broken down into fixed subtasks, each contributing to the final output. It is particularly useful in scenarios where step-by-step reasoning enhances accuracy.
    
    \textbf{Example Applications:}
    \begin{itemize}
        \item Generating marketing content in one language and then translating it into another while preserving nuances.
        \item Structuring document creation by first generating an outline, verifying its completeness, and then developing the full text.
    \end{itemize}

\subsection{Routing:Directing Inputs to Specialized Processes}

Routing \cite{anthropic2024agents, langgraph2025workflows} involves classifying an input and directing it to an appropriate specialized prompt or process. This method ensures distinct queries or tasks are handled separately, improving efficiency and response quality.
\begin{figure}[H]
    \centering
    \includegraphics[width=\linewidth]{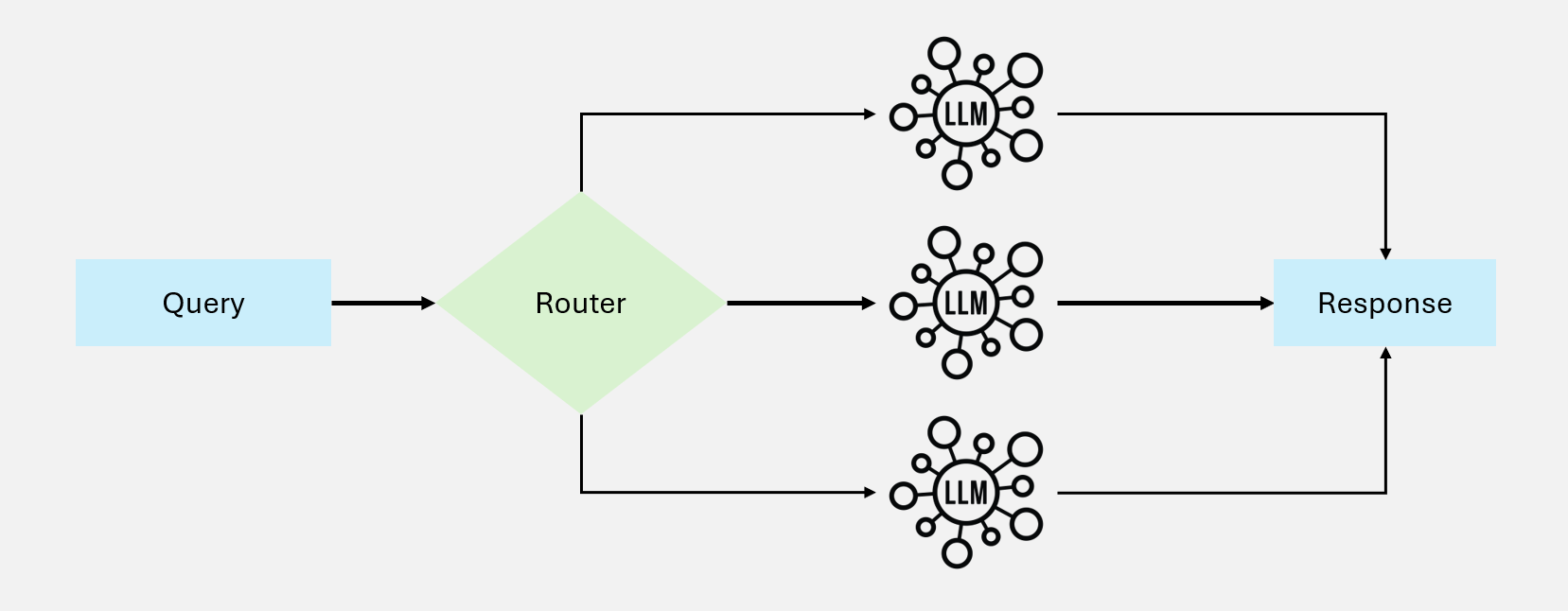}
    \caption{Illustration Routing Workflow}
    \label{fig:routing}
\end{figure}

    \textbf{When to Use:} Ideal for scenarios where different types of input require distinct handling strategies, ensuring optimized performance for each category.
    
    \textbf{Example Applications:}
    \begin{itemize}
        \item Directing customer service queries into categories such as technical support, refund requests, or general inquiries.
        \item Assigning simple queries to smaller models for cost efficiency, while complex requests go to advanced models.
    \end{itemize}

\subsection{Parallelization: Speeding Up Processing Through Concurrent Execution}

Parallelization \cite{anthropic2024agents, langgraph2025workflows} divides a task into independent processes that run simultaneously, reducing latency and improving throughput. It can be categorized into sectioning (independent subtasks) and voting (multiple outputs for accuracy).

\begin{figure}[h]
    \centering
    \includegraphics[width=\linewidth]{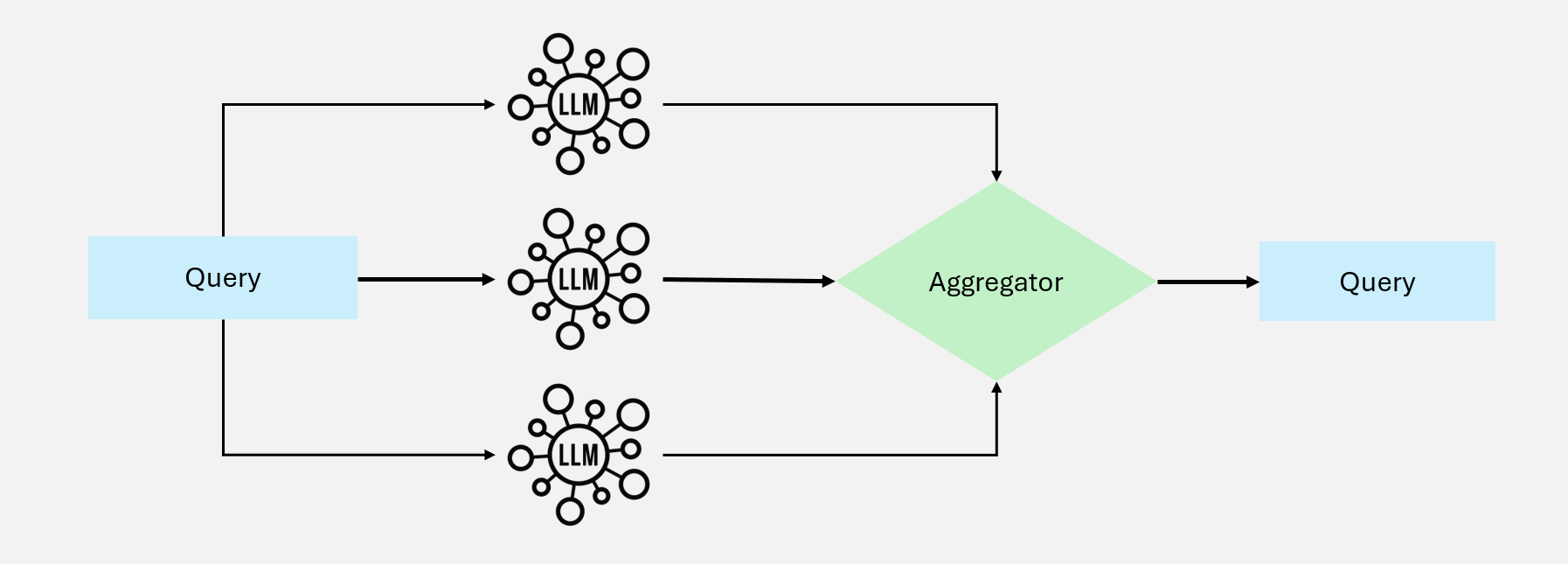}
    \caption{Illustration of Parallelization Workflow}
    \label{fig:parallelization}
\end{figure}

    \textbf{When to Use:} Useful when tasks can be executed independently to enhance speed or when multiple outputs improve confidence.
    
    \textbf{Example Applications:}
    \begin{itemize}
        \item \textbf{Sectioning:} Splitting tasks like content moderation, where one model screens input while another generates a response.
        \item \textbf{Voting:} Using multiple models to cross-check code for vulnerabilities or analyze content moderation decisions.
    \end{itemize}

\subsection{Orchestrator-Workers: Dynamic Task Delegation}

This workflow  \cite{anthropic2024agents, langgraph2025workflows} features a central orchestrator model that dynamically breaks tasks into subtasks, assigns them to specialized worker models, and compiles the results. Unlike parallelization, it adapts to varying input complexity.

\begin{figure}[h]
    \centering
    \includegraphics[width=\linewidth]{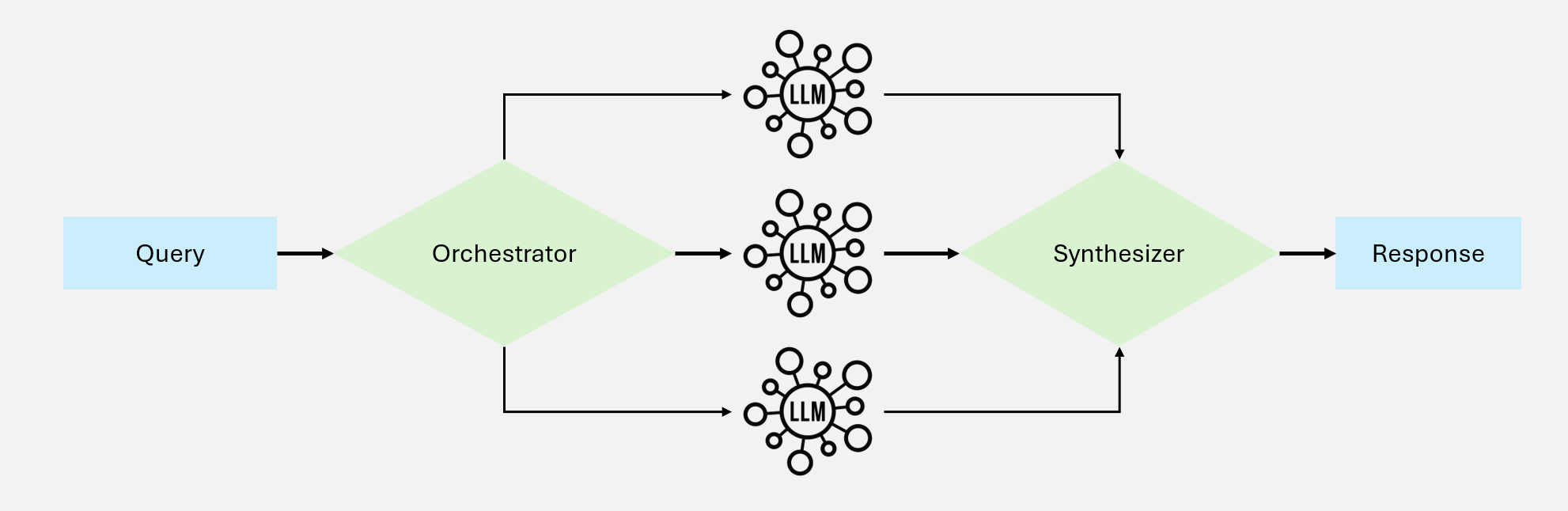}
    \caption{Illustration of Orchestrator-Workers Workflow}
    \label{fig:orchestrator_workers}
\end{figure}

    \textbf{When to Use:} Best suited for tasks requiring dynamic decomposition and real-time adaptation, where subtasks are not predefined.
    
    \textbf{Example Applications:}
    \begin{itemize}
        \item Automatically modifying multiple files in a codebase based on the nature of requested changes.
        \item Conducting real-time research by gathering and synthesizing relevant information from multiple sources.
    \end{itemize}

\subsection{Evaluator-Optimizer: Refining Output Through Iteration}

The evaluator-optimizer \cite{anthropic2024agents, langgraph2025workflows} workflow iteratively improves content by generating an initial output and refining it based on feedback from an evaluation model.

\begin{figure}[h]
    \centering
    \includegraphics[width=\linewidth]{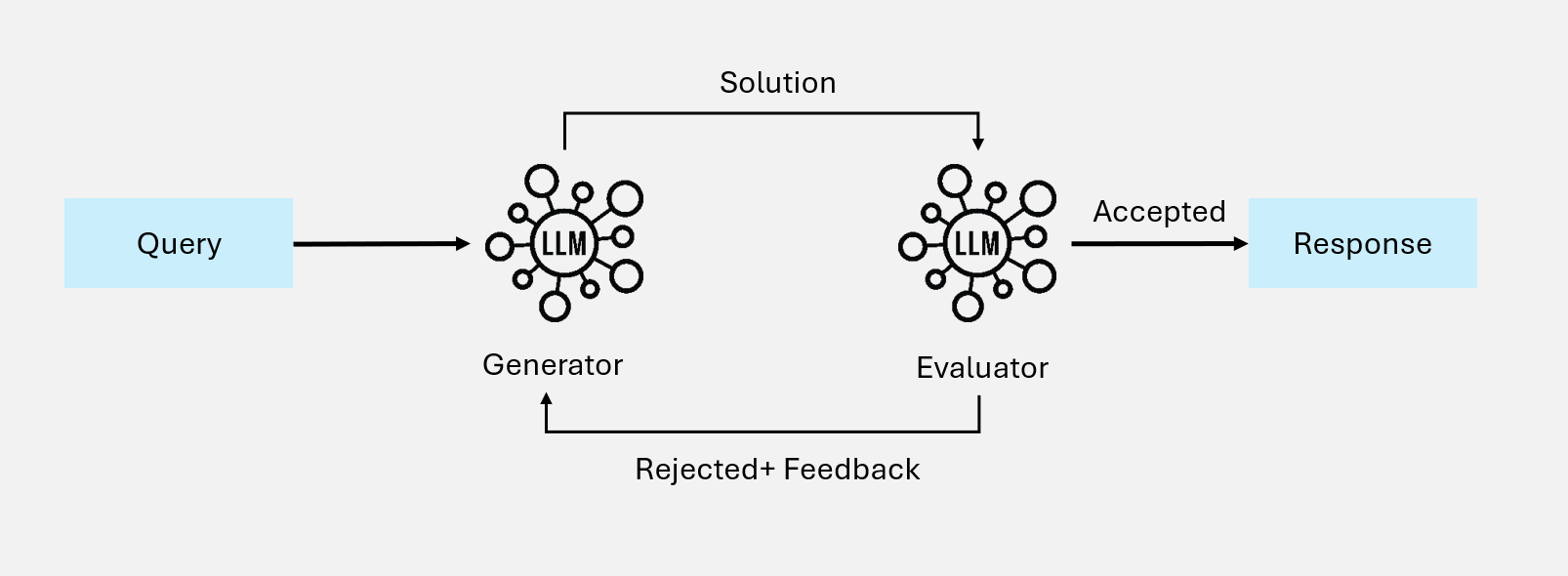}
    \caption{Illustration of Evaluator-Optimizer Workflow}
    \label{fig:evaluator_optimizer}
\end{figure}

    \textbf{When to Use:} Effective when iterative refinement significantly enhances response quality, especially when clear evaluation criteria exist.
    
    \textbf{Example Applications:}
    \begin{itemize}
        \item Improving literary translations through multiple evaluation and refinement cycles.
        \item Conducting multi-round research queries where additional iterations refine search results.
    \end{itemize}

\section{Taxonomy of Agentic RAG Systems}

Agentic Retrieval-Augmented Generation (RAG) systems can be categorized into distinct architectural frameworks based on their complexity and design principles. These include single-agent architectures, multi-agent systems, and hierarchical agentic architectures. Each framework is tailored to address specific challenges and optimize performance for diverse applications. This section provides a detailed taxonomy of these architectures, highlighting their characteristics, strengths, and limitations.

\subsection{Single-Agent Agentic RAG: Router}

\begin{figure}[htbp]
    \centering
    \includegraphics[width=\linewidth]{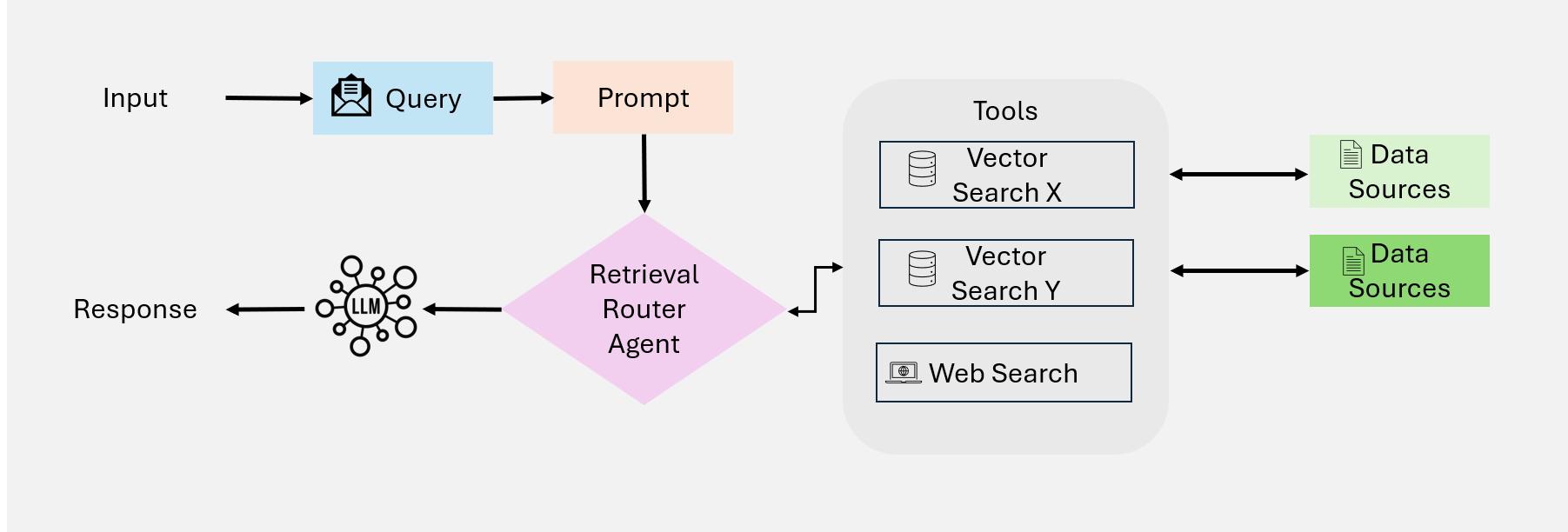}
    \caption{An Overview of Single Agentic RAG}
    \label{fig:single_agentic_RAG}
\end{figure}

A \textbf{Single-Agent Agentic RAG:} \cite{weaviate2025agenticrag} serves as a centralized decision-making system where a single agent manages the retrieval, routing, and integration of information (as shown in Figure. \ref{fig:single_agentic_RAG}). This architecture simplifies the system by consolidating these tasks into one unified agent, making it particularly effective for setups with a limited number of tools or data sources.
\paragraph{Workflow}
\begin{enumerate}
    \item \textbf{Query Submission and Evaluation:}
    The process begins when a user submits a query. A coordinating agent (or master retrieval agent) receives the query 
    and analyzes it to determine the most suitable sources of information.
    
    \item \textbf{Knowledge Source Selection:}
    Based on the query’s type, the coordinating agent chooses from a variety of retrieval options:
    \begin{itemize}
        \item \textbf{Structured Databases:} For queries requiring tabular data access, the system may 
        use a \textit{Text-to-SQL} engine that interacts with databases like PostgreSQL or MySQL.
        \item \textbf{Semantic Search:} When dealing with unstructured information, it retrieves 
        relevant documents (e.g., PDFs, books, organizational records) using vector-based retrieval.
        \item \textbf{Web Search:} For real-time or broad contextual information, the system 
        leverages a web search tool to access the latest online data.
        \item \textbf{Recommendation Systems:} For personalized or contextual queries, the system 
        taps into recommendation engines that provide tailored suggestions.
    \end{itemize}
    
    \item \textbf{Data Integration and LLM Synthesis:}
    Once the relevant data is retrieved from the chosen sources, it is passed to a 
    \textit{Large Language Model (LLM)}. The LLM synthesizes the gathered information, integrating 
    insights from multiple sources into a coherent and contextually relevant response.
    
    \item \textbf{Output Generation:}
    Finally, the system delivers a comprehensive, user-facing answer that addresses the original 
    query. This response is presented in an actionable, concise format and may optionally include 
    references or citations to the sources used.
\end{enumerate}

\paragraph{Key Features and Advantages.}
\begin{itemize}
    \item \textbf{Centralized Simplicity:} A single agent handles all retrieval and routing tasks, 
          making the architecture straightforward to design, implement, and maintain.
    \item \textbf{Efficiency \& Resource Optimization:} With fewer agents and simpler coordination, 
          the system demands fewer computational resources and can handle queries more quickly.
    \item \textbf{Dynamic Routing:} The agent evaluates each query in real-time, selecting 
          the most appropriate knowledge source (e.g., structured DB, semantic search, web search).
    \item \textbf{Versatility Across Tools:} Supports a variety of data sources and external APIs, 
          enabling both structured and unstructured workflows.
    \item \textbf{Ideal for Simpler Systems:} Suited for applications with well-defined tasks 
          or limited integration requirements (e.g., document retrieval, SQL-based workflows).
\end{itemize}

\begin{tcolorbox}[colframe=black, colback=white, boxrule=1pt, sharp corners=all, title=Use Case: Customer Support]
\textbf{Prompt:} Can you tell me the delivery status of my order?\\

\textbf{System Process (Single-Agent Workflow):}
\begin{enumerate}
    \item \textbf{Query Submission and Evaluation:}
    \begin{itemize}
        \item The user submits the query, which is received by the coordinating agent.
        \item The coordinating agent analyzes the query and determines the most appropriate sources of information.
    \end{itemize}
    \item \textbf{Knowledge Source Selection:}
    \begin{itemize}
        \item Retrieves tracking details from the order management database.
        \item Fetches real-time updates from the shipping provider’s API.
        \item Optionally conducts a web search to identify local conditions affecting delivery, such as weather or logistical delays.
    \end{itemize}
    \item \textbf{Data Integration and LLM Synthesis:}
    \begin{itemize}
        \item The relevant data is passed to the LLM, which synthesizes the information into a coherent response.
    \end{itemize}
    \item \textbf{Output Generation:}
    \begin{itemize}
        \item The system generates an actionable and concise response, providing live tracking updates and potential alternatives.
    \end{itemize}
\end{enumerate}

\textbf{Response:}\\
\textit{Integrated Response:} 
``Your package is currently in transit and expected to arrive tomorrow evening. The live tracking from UPS indicates it is at the regional distribution center.''
\end{tcolorbox}

\subsection{Multi-Agent Agentic RAG Systems:}

\begin{figure}[h]
    \centering
    \includegraphics[width=\linewidth]{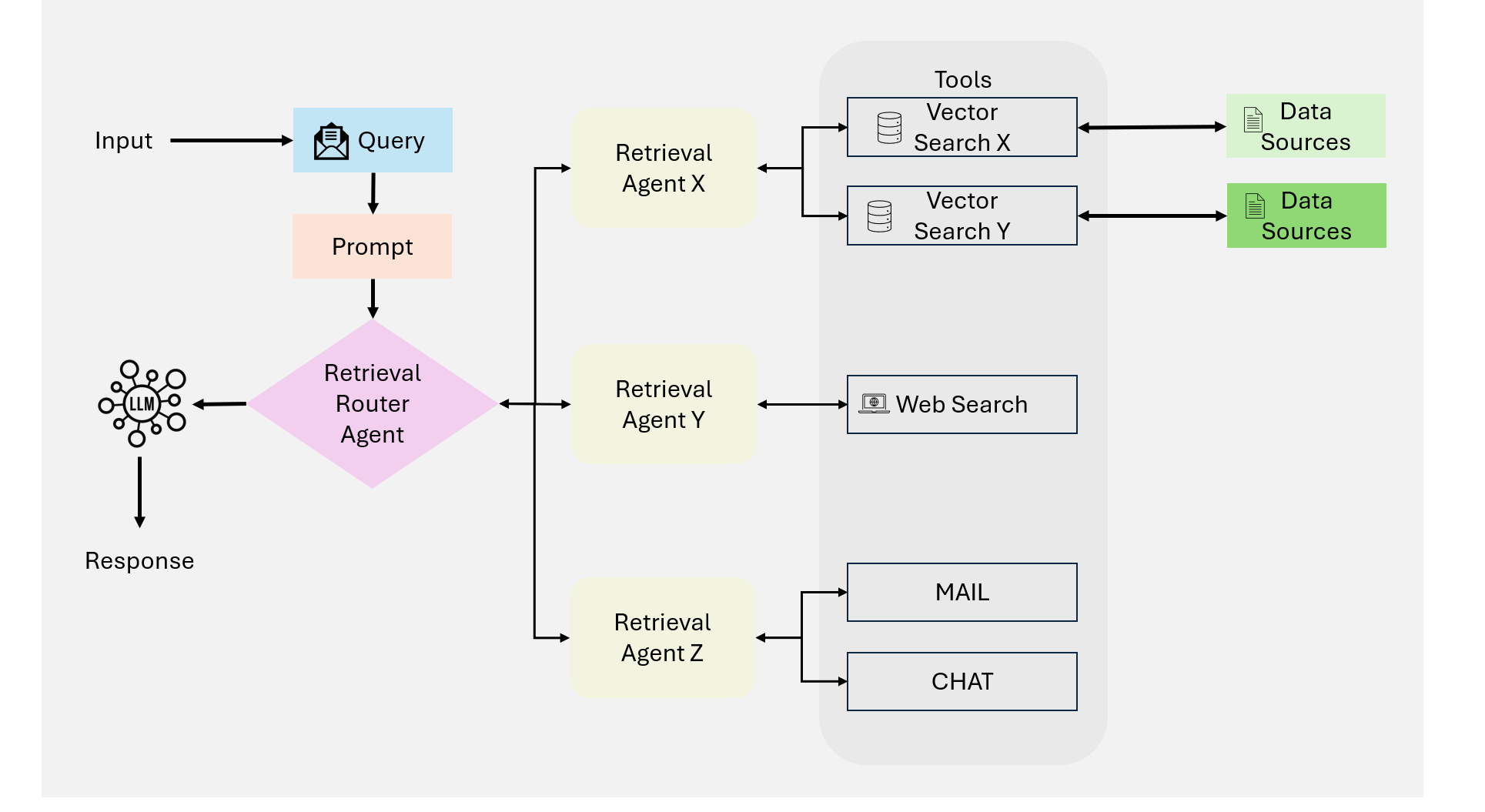}
    \caption{An Overview of Multi-Agent Agentic RAG Systems}
    \label{fig:multiagent_agentic_rag}
\end{figure}

\textbf{Multi-Agent RAG} \cite{weaviate2025agenticrag} represents a modular and scalable evolution of single-agent architectures, designed to handle complex workflows and diverse query types by leveraging multiple specialized agents (as shown in Figure \ref{fig:multiagent_agentic_rag}). Instead of relying on a single agent to manage all tasks—reasoning, retrieval, and response generation—this system distributes responsibilities across multiple agents, each optimized for a specific role or data source.

\paragraph{Workflow}
\begin{enumerate}
    \item \textbf{Query Submission}: The process begins with a user query, which is received by a \textit{coordinator agent} or master retrieval agent. This agent acts as the central orchestrator, delegating the query to specialized retrieval agents based on the query’s requirements.
    
    \item \textbf{Specialized Retrieval Agents}: The query is distributed among multiple retrieval agents, each focusing on a specific type of data source or task. Examples include:
    \begin{itemize}
        \item \textbf{Agent 1}: Handles structured queries, such as interacting with SQL-based databases like PostgreSQL or MySQL.
        \item \textbf{Agent 2}: Manages semantic searches for retrieving unstructured data from sources like PDFs, books, or internal records.
        \item \textbf{Agent 3}: Focuses on retrieving real-time public information from web searches or APIs.
        \item \textbf{Agent 4}: Specializes in recommendation systems, delivering context-aware suggestions based on user behavior or profiles.
    \end{itemize}
    
    \item \textbf{Tool Access and Data Retrieval}: Each agent routes the query to the appropriate tools or data sources within its domain, such as:
    \begin{itemize}
        \item \textit{Vector Search}: For semantic relevance.
        \item \textit{Text-to-SQL}: For structured data.
        \item \textit{Web Search}: For real-time public information.
        \item \textit{APIs}: For accessing external services or proprietary systems.
    \end{itemize}
    The retrieval process is executed in parallel, allowing for efficient processing of diverse query types.
    
    \item \textbf{Data Integration and LLM Synthesis}: Once retrieval is complete, the data from all agents is passed to a \textit{Large Language Model (LLM)}. The LLM synthesizes the retrieved information into a coherent and contextually relevant response, integrating insights from multiple sources seamlessly.
    
    \item \textbf{Output Generation}: The system generates a comprehensive response, which is delivered back to the user in an actionable and concise format.
\end{enumerate}




\paragraph{Key Features and Advantages.}
\begin{itemize}
    \item \textbf{Modularity}: Each agent operates independently, allowing for seamless addition or removal of agents based on system requirements.
    \item \textbf{Scalability}: Parallel processing by multiple agents enables the system to handle high query volumes efficiently.
    \item \textbf{Task Specialization}: Each agent is optimized for a specific type of query or data source, improving accuracy and retrieval relevance.
    \item \textbf{Efficiency}: By distributing tasks across specialized agents, the system minimizes bottlenecks and enhances performance for complex workflows.
    \item \textbf{Versatility}: Suitable for applications spanning multiple domains, including research, analytics, decision-making, and customer support.
\end{itemize}

\paragraph{Challenges}
\begin{itemize}
    \item \textbf{Coordination Complexity}: Managing inter-agent communication and task delegation requires sophisticated orchestration mechanisms.
    \item \textbf{Computational Overhead}: Parallel processing of multiple agents can increase resource usage.
    \item \textbf{Data Integration}: Synthesizing outputs from diverse sources into a cohesive response is non-trivial and requires advanced LLM capabilities.
\end{itemize}

\begin{tcolorbox}[colframe=black, colback=white, boxrule=1pt, sharp corners=all, title=Use Case: Multi-Domain Research Assistant]
\textbf{Prompt:} What are the economic and environmental impacts of renewable energy adoption in Europe?\\[0.2cm]

\textbf{System Process (Multi-Agent Workflow):}
\begin{itemize}
    \item \textbf{Agent 1:} Retrieves statistical data from economic databases using SQL-based queries.
    \item \textbf{Agent 2:} Searches for relevant academic papers using semantic search tools.
    \item \textbf{Agent 3:} Performs a web search for recent news and policy updates on renewable energy.
    \item \textbf{Agent 4:} Consults a recommendation system to suggest related content, such as reports or expert commentary.
\end{itemize}

\textbf{Response:}\\
\textit{Integrated Response:} 
``Adopting renewable energy in Europe has led to a 20\% reduction in greenhouse gas emissions over the past 
decade, according to EU policy reports. Economically, renewable energy investments have generated approximately 
1.2 million jobs, with significant growth in solar and wind sectors. Recent academic studies also highlight potential 
trade-offs in grid stability and energy storage costs.''
\end{tcolorbox}

\subsection{Hierarchical Agentic RAG Systems}

\begin{figure}[h]
    \centering
    \includegraphics[width=0.9\linewidth]{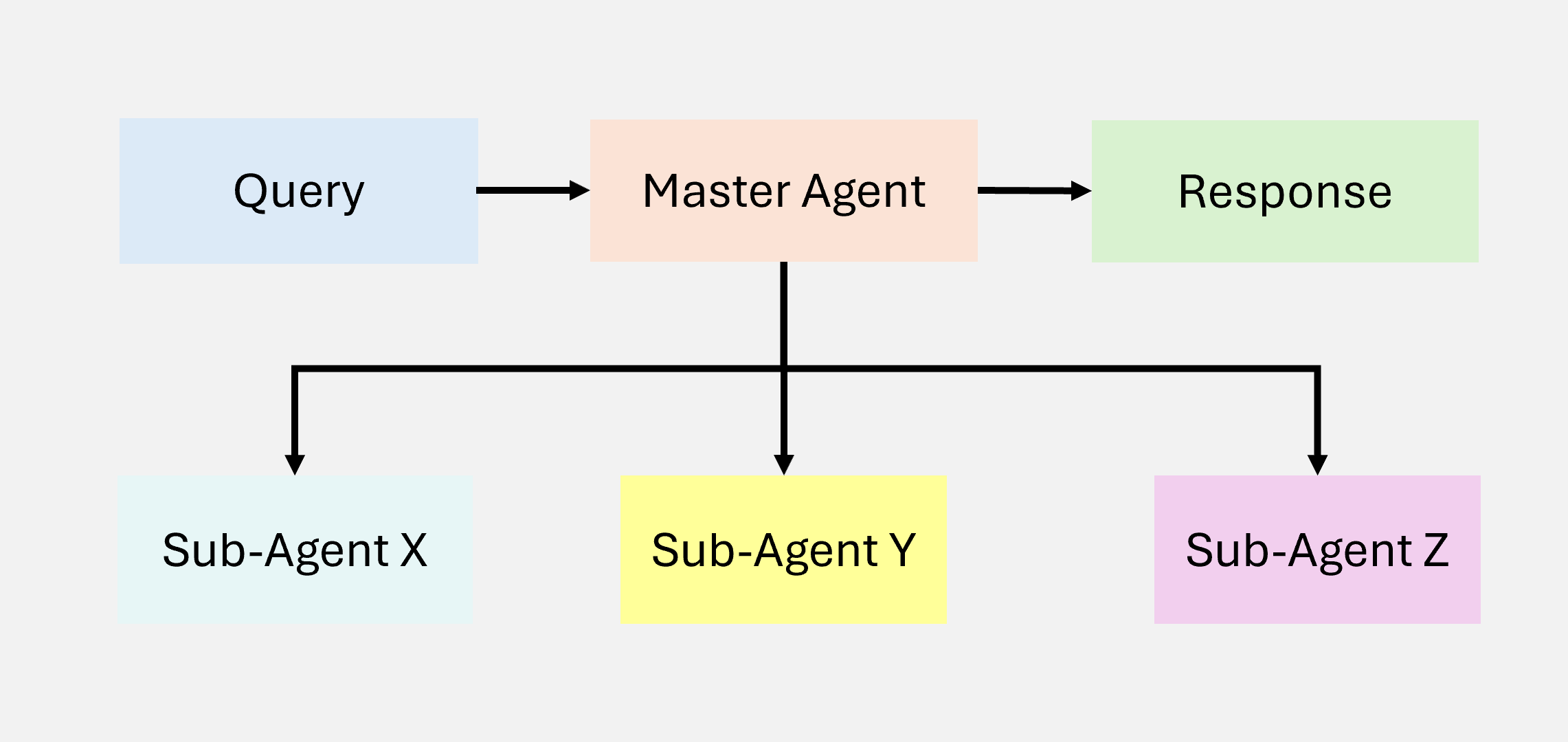}
    \caption{An illustration of Hierarchical Agentic RAG}
    \label{fig:hierarchical_agentic_rag}
\end{figure}

\textbf{Hierarchical Agentic RAG:} \cite{ravuru2024agenticrag} systems employ a structured, multi-tiered approach to information retrieval 
and processing, enhancing both efficiency and strategic decision-making as shown in Figure \ref{fig:hierarchical_agentic_rag}. Agents are organized in a hierarchy, 
with higher-level agents overseeing and directing lower-level agents. This structure enables multi-level 
decision-making, ensuring that queries are handled by the most appropriate resources.

\paragraph{Workflow}
\begin{enumerate}
    \item \textbf{Query Reception}: A user submits a query, received by a \textit{top-tier agent} responsible 
    for initial assessment and delegation.
    
    \item \textbf{Strategic Decision-Making}: The top-tier agent evaluates the query’s complexity and decides 
    which subordinate agents or data sources to prioritize. Certain databases, APIs, or retrieval tools may be 
    deemed more reliable or relevant based on the query’s domain.

    \item \textbf{Delegation to Subordinate Agents}: The top-tier agent assigns tasks to lower-level agents 
    specialized in particular retrieval methods (e.g., SQL databases, web search, or proprietary systems). 
    These agents execute their assigned tasks independently.
    
    \item \textbf{Aggregation and Synthesis}: The results from subordinate agents are collected and integrated 
    by the higher-level agent, which synthesizes the information into a coherent response.
    
    \item \textbf{Response Delivery}: The final, synthesized answer is returned to the user, ensuring that 
    the response is both comprehensive and contextually relevant.
\end{enumerate}

\paragraph{Key Features and Advantages.}
\begin{itemize}
    \item \textbf{Strategic Prioritization}: Top-tier agents can prioritize data sources or tasks based on 
    query complexity, reliability, or context.
    \item \textbf{Scalability}: Distributing tasks across multiple agent tiers enables handling of highly 
    complex or multi-faceted queries.
    \item \textbf{Enhanced Decision-Making}: Higher-level agents apply strategic oversight to improve overall 
    accuracy and coherence of responses.
\end{itemize}

\paragraph{Challenges}
\begin{itemize}
    \item \textbf{Coordination Complexity}: Maintaining robust inter-agent communication across multiple levels 
    can increase orchestration overhead.
    \item \textbf{Resource Allocation}: Efficiently distributing tasks among tiers to avoid bottlenecks is 
    non-trivial.
\end{itemize}

\begin{tcolorbox}[colframe=black, colback=white, boxrule=1pt, sharp corners=all, title=Use Case: Financial Analysis System]
\textbf{Prompt:} What are the best investment options given the current market trends in renewable energy?\\[0.2cm]

\textbf{System Process (Hierarchical Agentic Workflow):}
\begin{enumerate}
    \item \textbf{Top-Tier Agent}: Assesses the query’s complexity and prioritizes reliable financial databases 
    and economic indicators over less validated data sources.
    \item \textbf{Mid-Level Agent}: Retrieves real-time market data (e.g., stock prices, sector performance) 
    from proprietary APIs and structured SQL databases.
    \item \textbf{Lower-Level Agent(s)}: Conducts web searches for recent policy announcements and 
    consults recommendation systems that track expert opinions and news analytics.
    \item \textbf{Aggregation and Synthesis}: The top-tier agent compiles the results, integrating 
    quantitative data with policy insights.
\end{enumerate}

\textbf{Response:}\\
\textit{Integrated Response:} 
``Based on current market data, renewable energy stocks have shown a 15\% growth over the past quarter, 
driven by supportive government policies and heightened investor interest. Analysts suggest that wind 
and solar sectors, in particular, may experience continued momentum, while emerging technologies 
like green hydrogen present moderate risk but potentially high returns.''
\end{tcolorbox}

\subsection{Agentic Corrective RAG}

\begin{figure}[h]
    \centering
    \includegraphics[width=\linewidth]{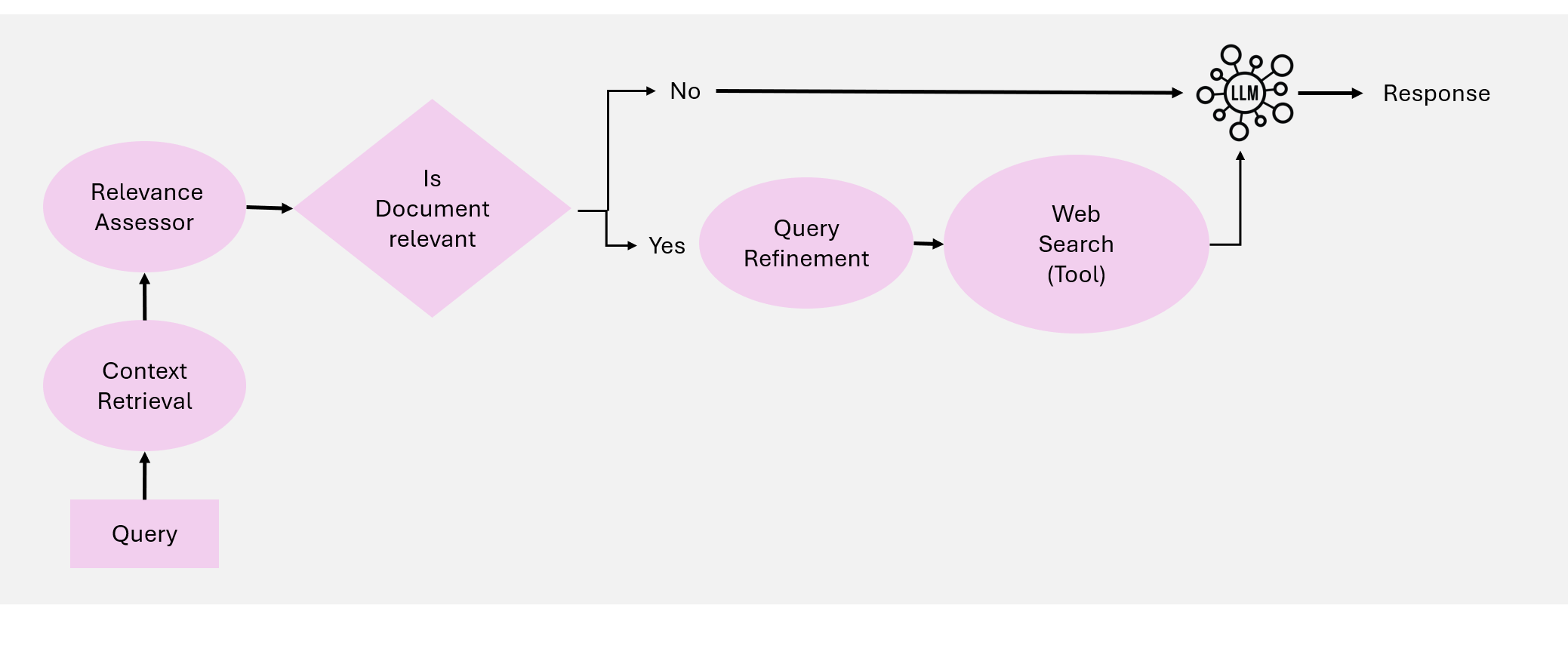}
    \caption{Overview of Agentic Corrective RAG}
    \label{fig:agentic_corrective_rag}
\end{figure}

\textbf{Corrective RAG :} introduces mechanisms to self-correct retrieval results, enhancing document utilization and improving response generation quality as demonstrated in Figure \ref{fig:agentic_corrective_rag}. By embedding intelligent agents into the workflow, Corrective RAG \cite{yan2024correctiveretrievalaugmentedgeneration} \cite{langgraph_crag_tutorial} ensures iterative refinement of context documents and responses, minimizing errors and maximizing relevance.

\paragraph{Key Idea of Corrective RAG:}
The core principle of Corrective RAG lies in its ability to evaluate retrieved documents dynamically, perform corrective actions, and refine queries to enhance the quality of generated responses. Corrective RAG adjusts its approach as follows:
\begin{itemize}
    \item \textbf{Document Relevance Evaluation:} Retrieved documents are assessed for relevance by the \textit{Relevance Evaluation Agent}. Documents below the relevance threshold trigger corrective steps.
    \item \textbf{Query Refinement and Augmentation:} Queries are refined by the \textit{Query Refinement Agent}, which leverages semantic understanding to optimize retrieval for better results.
    \item \textbf{Dynamic Retrieval from External Sources:} When context is insufficient, the \textit{External Knowledge Retrieval Agent} performs web searches or accesses alternative data sources to supplement the retrieved documents.
    \item \textbf{Response Synthesis:} All validated and refined information is passed to the \textit{Response Synthesis Agent} for final response generation.
\end{itemize}

\paragraph{Workflow:}The Corrective RAG system is built on five key agents:
\begin{enumerate}
    \item \textbf{Context Retrieval Agent:} Responsible for retrieving initial context documents from a vector database.
    \item \textbf{Relevance Evaluation Agent:} Assesses the retrieved documents for relevance and flags any irrelevant or ambiguous documents for corrective actions.
    \item \textbf{Query Refinement Agent:} Rewrites queries to improve retrieval, leveraging semantic understanding to optimize results.
    \item \textbf{External Knowledge Retrieval Agent:} Performs web searches or accesses alternative data sources when the context documents are insufficient.
    \item \textbf{Response Synthesis Agent:} Synthesizes all validated information into a coherent and accurate response.
\end{enumerate}

\paragraph{Key Features and Advantages:}
\begin{itemize}
    \item \textbf{Iterative Correction:} Ensures high response accuracy by dynamically identifying and correcting irrelevant or ambiguous retrieval results.
    \item \textbf{Dynamic Adaptability:} Incorporates real-time web searches and query refinement for enhanced retrieval precision.
    \item \textbf{Agentic Modularity:} Each agent performs specialized tasks, ensuring efficient and scalable operation.
    \item \textbf{Factuality Assurance:} By validating all retrieved and generated content, Corrective RAG minimizes the risk of hallucination or misinformation.
\end{itemize}

\begin{tcolorbox}[colframe=black, colback=white, boxrule=1pt, sharp corners=all, title=Use Case: Academic Research Assistant]
\textbf{Prompt:} What are the latest findings in generative AI research?\\[0.2cm]

\textbf{System Process (Corrective RAG Workflow):}
\begin{enumerate}
    \item \textbf{Query Submission:} A user submits the query to the system.
    \item \textbf{Context Retrieval:} 
    \begin{itemize}
        \item The \textit{Context Retrieval Agent} retrieves initial documents from a database of published papers on generative AI.
        \item The retrieved documents are passed to the next step for evaluation.
    \end{itemize}
    \item \textbf{Relevance Evaluation:} 
    \begin{itemize}
        \item The \textit{Relevance Evaluation Agent} assesses the documents for alignment with the query.
        \item Documents are classified into relevant, ambiguous, or irrelevant categories. Irrelevant documents are flagged for corrective actions.
    \end{itemize}
    \item \textbf{Corrective Actions (if needed):}
    \begin{itemize}
        \item The \textit{Query Refinement Agent} rewrites the query to improve specificity and relevance.
        \item The \textit{External Knowledge Retrieval Agent} performs web searches to fetch additional papers and reports from external sources.
    \end{itemize}
    \item \textbf{Response Synthesis:}
    \begin{itemize}
        \item The \textit{Response Synthesis Agent} integrates validated documents into a coherent and comprehensive summary.
    \end{itemize}
\end{enumerate}

\textbf{Response:}\\
\textit{Integrated Response:} 
``Recent findings in generative AI highlight advancements in diffusion models, reinforcement learning for text-to-video tasks, and optimization techniques for large-scale model training. For more details, refer to studies published in NeurIPS 2024 and AAAI 2025.''
\end{tcolorbox}
\subsection{Adaptive Agentic RAG}
\begin{figure}[h]
    \centering
    \includegraphics[width=\linewidth]{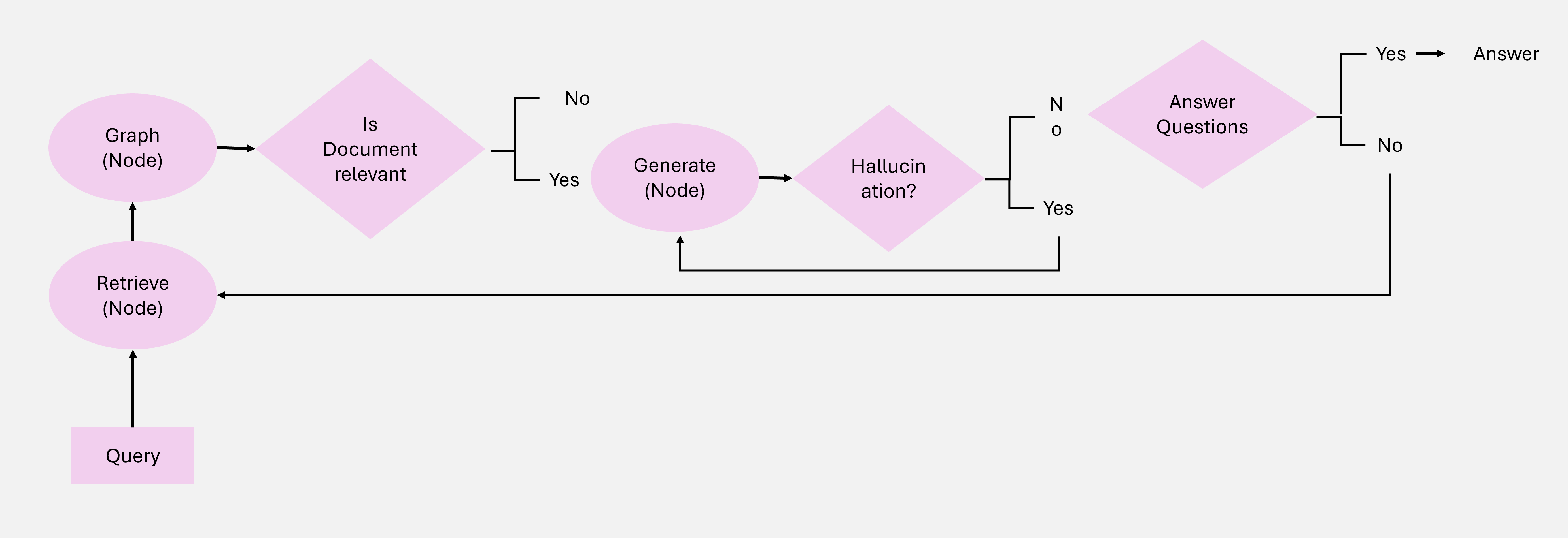}
    \caption{An Overview of Adaptive Agentic RAG}
    \label{fig:adaptive_rag_agent}
\end{figure}

\textbf{Adaptive Retrieval-Augmented Generation (Adaptive RAG)} \cite{jeong2024adaptiveraglearningadaptrag} enhances the flexibility and efficiency of large language models (LLMs) by dynamically adjusting query handling strategies based on the complexity of the incoming query. Unlike static retrieval workflows, Adaptive RAG \cite{langgraph_adaptive_rag_tutorial} employs a classifier to assess query complexity and determine the most appropriate approach, ranging from single-step retrieval to multi-step reasoning, or even bypassing retrieval altogether for straightforward queries as illustrated in Figure \ref{fig:adaptive_rag_agent}.

\paragraph{Key Idea of Adaptive RAG}
The core principle of Adaptive RAG lies in its ability to dynamically tailor retrieval strategies based on the complexity of the query. Adaptive RAG adjusts its approach as follows:

\begin{itemize}
    \item \textbf{Straightforward Queries:} For fact-based questions that require no additional retrieval (e.g., \textit{"What is the boiling point of water?"}), the system directly generates an answer using pre-existing knowledge.
    \item \textbf{Simple Queries:} For moderately complex tasks requiring minimal context (e.g., \textit{"What is the status of my latest electricity bill?"}), the system performs a single-step retrieval to fetch the relevant details.
    \item \textbf{Complex Queries:} For multi-layered queries requiring iterative reasoning (e.g., \textit{"How has the population of City X changed over the past decade, and what are the contributing factors?"}), the system employs multi-step retrieval, progressively refining intermediate results to provide a comprehensive answer.
\end{itemize}

\paragraph{Workflow:}The Adaptive RAG system is built on three primary components:
\begin{enumerate}
    \item \textbf{Classifier Role:} 
    \begin{itemize}
        \item A smaller language model analyzes the query to predict its complexity.
        \item The classifier is trained using automatically labeled datasets, derived from past model outcomes and query patterns.
    \end{itemize}
    \item \textbf{Dynamic Strategy Selection:}
    \begin{itemize}
        \item For straightforward queries, the system avoids unnecessary retrieval, directly leveraging the LLM for response generation.
        \item For simple queries, it employs a single-step retrieval process to fetch relevant context.
        \item For complex queries, it activates multi-step retrieval to ensure iterative refinement and enhanced reasoning.
    \end{itemize}
    \item \textbf{LLM Integration:}
    \begin{itemize}
        \item The LLM synthesizes retrieved information into a coherent response.
        \item Iterative interactions between the LLM and the classifier enable refinement for complex queries.
    \end{itemize}
\end{enumerate}

\paragraph{Key Features and Advantages}
\begin{itemize}
    \item \textbf{Dynamic Adaptability:} Adjusts retrieval strategies based on query complexity, optimizing both computational efficiency and response accuracy.
    \item \textbf{Resource Efficiency:} Minimizes unnecessary overhead for simple queries while ensuring thorough processing for complex ones.
    \item \textbf{Enhanced Accuracy:} Iterative refinement ensures that complex queries are resolved with high precision.
    \item \textbf{Flexibility:} Can be extended to incorporate additional pathways, such as domain-specific tools or external APIs.
\end{itemize}

\begin{tcolorbox}[colframe=black, colback=white, boxrule=1pt, sharp corners=all, title=Use Case: Customer Support Assistant]
\textbf{Prompt:} Why is my package delayed, and what alternatives do I have?\\[0.2cm]

\textbf{System Process (Adaptive RAG Workflow):}
\begin{enumerate}
    \item \textbf{Query Classification:}
    \begin{itemize}
        \item The classifier analyzes the query and determines it to be complex, requiring multi-step reasoning.
    \end{itemize}
    \item \textbf{Dynamic Strategy Selection:} 
    \begin{itemize}
        \item The system activates a multi-step retrieval process based on the complexity classification.
    \end{itemize}
    \item \textbf{Multi-Step Retrieval:}
    \begin{itemize}
        \item Retrieves tracking details from the order database.
        \item Fetches real-time status updates from the shipping provider API.
        \item Conducts a web search for external factors such as weather conditions or local disruptions.
    \end{itemize}
    \item \textbf{Response Synthesis:} 
    \begin{itemize}
        \item The LLM integrates all retrieved information, synthesizing a comprehensive and actionable response.
    \end{itemize}
\end{enumerate}

\textbf{Response:}\\
\textit{Integrated Response:} 
``Your package is delayed due to severe weather conditions in your region. It is currently at the local distribution center and will be delivered in 2 days. Alternatively, you may opt for a local pickup from the facility.''
\end{tcolorbox}

\subsection{Graph-Based Agentic RAG}

\subsubsection{Agent-G: Agentic Framework for Graph RAG}

\begin{figure}[h]
    \centering
    \includegraphics[width=0.8\linewidth]{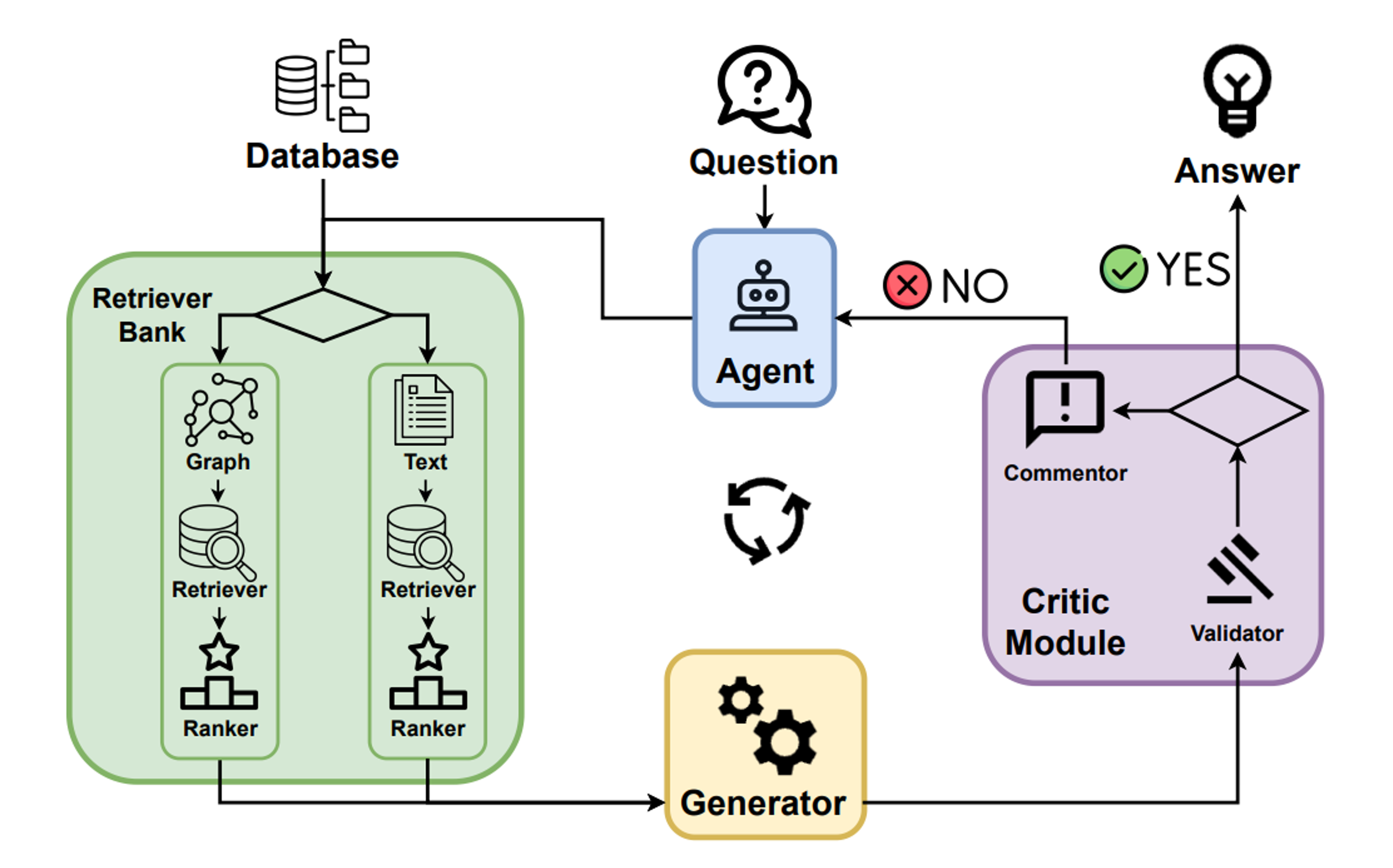}
    \caption{An Overview of Agent-G: Agentic Framework for Graph RAG \cite{lee2024agentg}}
    \label{fig:agent_g}
\end{figure}

\textbf{Agent-G} \cite{lee2024agentg}: introduces a novel agentic architecture that integrates graph knowledge bases with unstructured document retrieval. By combining structured and unstructured data sources, this framework enhances retrieval-augmented generation (RAG) systems with improved reasoning and retrieval accuracy. It employs modular retriever banks, dynamic agent interaction, and feedback loops to ensure high-quality outputs as shown in Figure \ref{fig:agent_g}.

\paragraph{Key Idea of Agent-G}
The core principle of Agent-G lies in its ability to dynamically assign retrieval tasks to specialized agents, leveraging both graph knowledge bases and textual documents. Agent-G adjusts its retrieval strategy as follows:
\begin{itemize}
    \item \textbf{Graph Knowledge Bases:} Structured data is used to extract relationships, hierarchies, and connections (e.g., disease-to-symptom mappings in healthcare).
    \item \textbf{Unstructured Documents:} Traditional text retrieval systems provide contextual information to complement graph data.
    \item \textbf{Critic Module:} Evaluates the relevance and quality of retrieved information, ensuring alignment with the query.
    \item \textbf{Feedback Loops:} Refines retrieval and synthesis through iterative validation and re-querying.
\end{itemize}

\paragraph{Workflow:} The Agent-G system is built on four primary components:

\begin{enumerate}
    \item \textbf{Retriever Bank:} 
    \begin{itemize}
        \item A modular set of agents specializes in retrieving graph-based or unstructured data.
        \item Agents dynamically select relevant sources based on the query’s requirements.
    \end{itemize}
    \item \textbf{Critic Module:}
    \begin{itemize}
        \item Validates retrieved data for relevance and quality.
        \item Flags low-confidence results for re-retrieval or refinement.
    \end{itemize}
    \item \textbf{Dynamic Agent Interaction:}
    \begin{itemize}
        \item Task-specific agents collaborate to integrate diverse data types.
        \item Ensures cohesive retrieval and synthesis across graph and text sources.
    \end{itemize}
    \item \textbf{LLM Integration:}
    \begin{itemize}
        \item Synthesizes validated data into a coherent response.
        \item Iterative feedback from the critic ensures alignment with the query’s intent.
    \end{itemize}
\end{enumerate}

\paragraph{Key Features and Advantages}
\begin{itemize}
    \item \textbf{Enhanced Reasoning:} Combines structured relationships from graphs with contextual information from unstructured documents.
    \item \textbf{Dynamic Adaptability:} Adjusts retrieval strategies dynamically based on query requirements.
    \item \textbf{Improved Accuracy:} Critic module reduces the risk of irrelevant or low-quality data in responses.
    \item \textbf{Scalable Modularity:} Supports the addition of new agents for specialized tasks, enhancing scalability.
\end{itemize}

\begin{tcolorbox}[colframe=black, colback=white, boxrule=1pt, sharp corners=all, title=Use Case: Healthcare Diagnostics]
\textbf{Prompt:} What are the common symptoms of Type 2 Diabetes, and how are they related to heart disease?\\[0.2cm]

\textbf{System Process (Agent-G Workflow):}
\begin{enumerate}
    \item \textbf{Query Reception and Assignment:} The system receives the query and identifies the need for both graph-structured and unstructured data to answer the question comprehensively.
    \item \textbf{Graph Retriever:} 
    \begin{itemize}
        \item Extracts relationships between Type 2 Diabetes and heart disease from a medical knowledge graph.
        \item Identifies shared risk factors such as obesity and high blood pressure by exploring graph hierarchies and relationships.
    \end{itemize}
    \item \textbf{Document Retriever:}
    \begin{itemize}
        \item Retrieves descriptions of Type 2 Diabetes symptoms (e.g., increased thirst, frequent urination, fatigue) from medical literature.
        \item Adds contextual information to complement the graph-based insights.
    \end{itemize}
    \item \textbf{Critic Module:}
    \begin{itemize}
        \item Evaluates the relevance and quality of the retrieved graph data and document data.
        \item Flags low-confidence results for refinement or re-querying.
    \end{itemize}
    \item \textbf{Response Synthesis:} The LLM integrates validated data from the Graph Retriever and Document Retriever into a coherent response, ensuring alignment with the query’s intent.
\end{enumerate}

\textbf{Response:}\\
\textit{Integrated Response:} 
``Type 2 Diabetes symptoms include increased thirst, frequent urination, and fatigue. Studies show a 50\% correlation between diabetes and heart disease, primarily through shared risk factors such as obesity and high blood pressure.''
\end{tcolorbox}

\subsubsection{GeAR: Graph-Enhanced Agent for Retrieval-Augmented Generation}

\begin{figure}[h]
    \centering
    \includegraphics[width=0.8\linewidth]{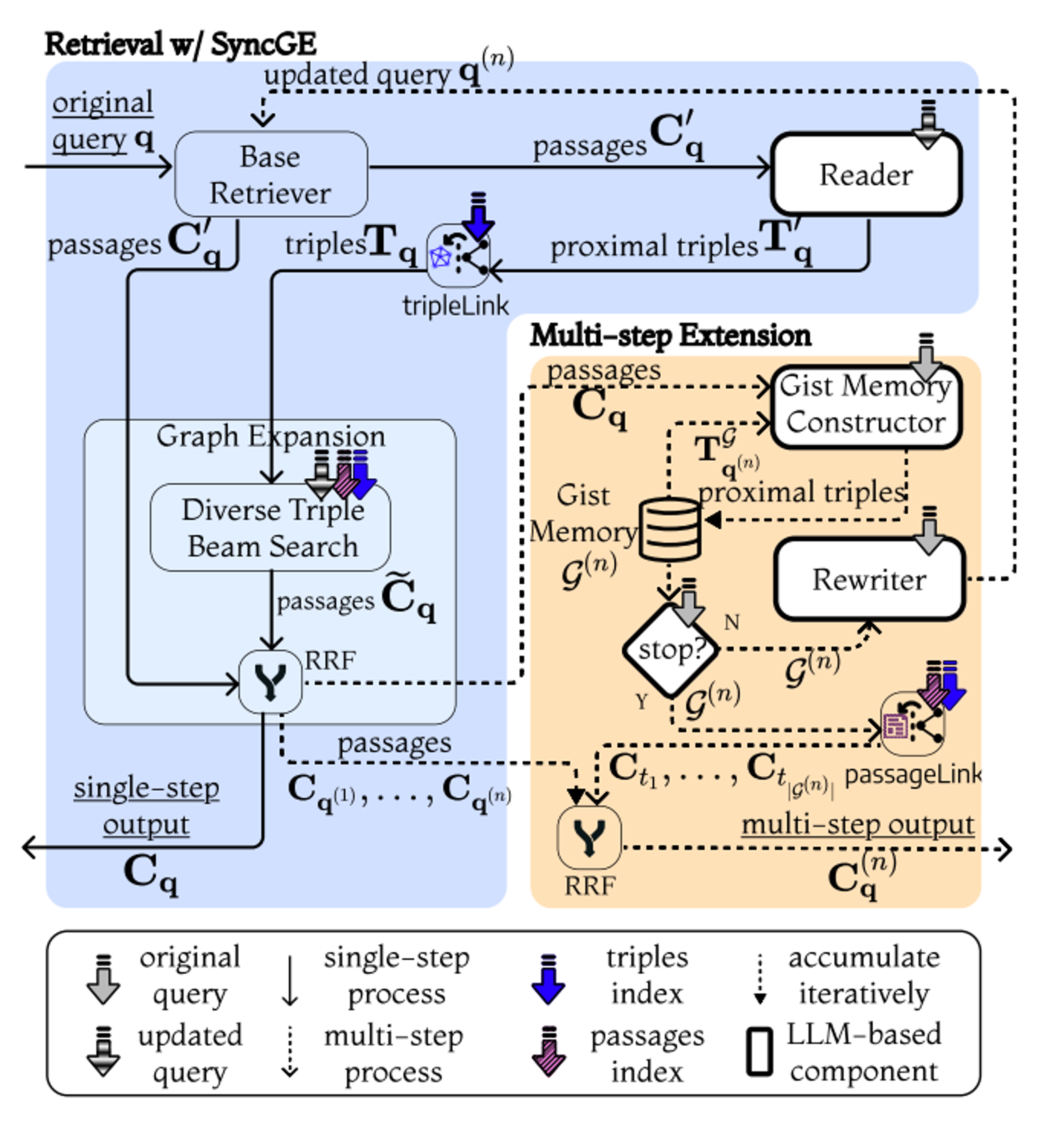}
    \caption{An Overview of GeAR: Graph-Enhanced Agent for Retrieval-Augmented Generation\cite{shen2024geargraphenhancedagentretrievalaugmented}}
    \label{fig:gear_overview}
\end{figure}

\textbf{GeAR} \cite{shen2024geargraphenhancedagentretrievalaugmented}: introduces an agentic framework that enhances traditional Retrieval-Augmented Generation (RAG) systems by incorporating graph-based retrieval mechanisms. By leveraging graph expansion techniques and an agent-based architecture, GeAR addresses challenges in multi-hop retrieval scenarios, improving the system's ability to handle complex queries as shown in Figure \ref{fig:gear_overview}.

\paragraph{Key Idea of GeAR}
GeAR advances RAG performance through two primary innovations:
\begin{itemize}
    \item \textbf{Graph Expansion:} Enhances conventional base retrievers (e.g., BM25) by expanding the retrieval process to include graph-structured data, enabling the system to capture complex relationships and dependencies between entities.
    \item \textbf{Agent Framework:} Incorporates an agent-based architecture that utilizes graph expansion to manage retrieval tasks more effectively, allowing for dynamic and autonomous decision-making in the retrieval process.
\end{itemize}

\paragraph{Workflow:}The GeAR system operates through the following components:
\begin{enumerate}
    \item \textbf{Graph Expansion Module:}
    \begin{itemize}
        \item Integrates graph-based data into the retrieval process, allowing the system to consider relationships between entities during retrieval.
        \item Enhances the base retriever's ability to handle multi-hop queries by expanding the search space to include connected entities.
    \end{itemize}
    \item \textbf{Agent-Based Retrieval:}
    \begin{itemize}
        \item Employs an agent framework to manage the retrieval process, enabling dynamic selection and combination of retrieval strategies based on the query's complexity.
        \item Agents can autonomously decide to utilize graph-expanded retrieval paths to improve the relevance and accuracy of retrieved information.
    \end{itemize}
    \item \textbf{LLM Integration:}
    \begin{itemize}
        \item Combines the retrieved information, enriched by graph expansion, with the capabilities of a Large Language Model (LLM) to generate coherent and contextually relevant responses.
        \item The integration ensures that the generative process is informed by both unstructured documents and structured graph data.
    \end{itemize}
\end{enumerate}

\paragraph{Key Features and Advantages}
\begin{itemize}
    \item \textbf{Enhanced Multi-Hop Retrieval:} GeAR's graph expansion allows the system to handle complex queries that require reasoning over multiple interconnected pieces of information.
    \item \textbf{Agentic Decision-Making:} The agent framework enables dynamic and autonomous selection of retrieval strategies, improving efficiency and relevance.
    \item \textbf{Improved Accuracy:} By incorporating structured graph data, GeAR enhances the precision of retrieved information, leading to more accurate and contextually appropriate responses.
    \item \textbf{Scalability:} The modular nature of the agent framework allows for the integration of additional retrieval strategies and data sources as needed.
\end{itemize}

\begin{tcolorbox}[colframe=black, colback=white, boxrule=1pt, sharp corners=all, title=Use Case: Multi-Hop Question Answering]
\textbf{Prompt:} Which author influenced the mentor of J.K. Rowling?\\[0.2cm]

\textbf{System Process (GeAR Workflow):}
\begin{enumerate}
    \item \textbf{Top-Tier Agent}: Evaluates the query’s multi-hop nature and determines that a combination of graph expansion and document retrieval is necessary to answer the question.
    \item \textbf{Graph Expansion Module}: 
    \begin{itemize}
        \item Identifies that J.K. Rowling’s mentor is a key entity in the query.
        \item Traces the literary influences on that mentor by exploring graph-structured data on literary relationships.
    \end{itemize}
    \item \textbf{Agent-Based Retrieval}: 
    \begin{itemize}
        \item An agent autonomously selects the graph-expanded retrieval path to gather relevant information about the mentor’s influences.
        \item Integrates additional context by querying textual data sources for unstructured details about the mentor and their influences.
    \end{itemize}
    \item \textbf{Response Synthesis}: Combines insights from the graph and document retrieval processes using the LLM to generate a response that accurately reflects the complex relationships in the query.
\end{enumerate}

\textbf{Response:}\\
\textit{Integrated Response:} 
``J.K. Rowling’s mentor, [Mentor Name], was heavily influenced by [Author Name], known for their [notable works or genre]. This connection highlights the layered relationships in literary history, where influential ideas often pass through multiple generations of authors.''
\end{tcolorbox}

\subsection{Agentic Document Workflows in Agentic RAG}

\begin{figure}[h]
    \centering
    \includegraphics[width=\linewidth]{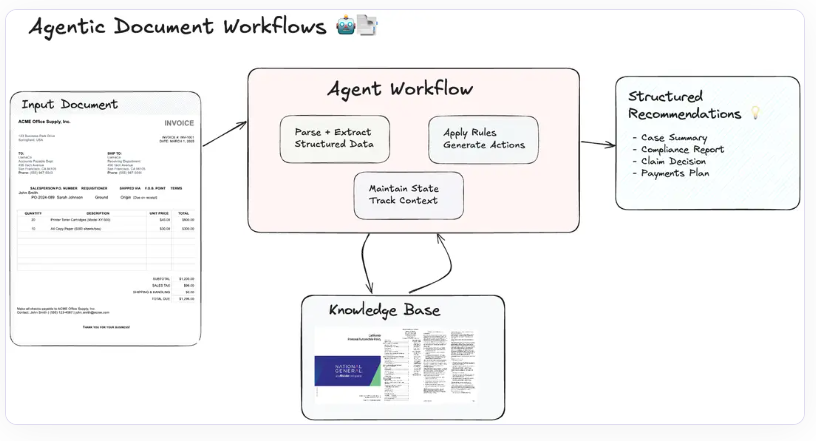}
    \caption{An Overview of Agentic Document Workflows (ADW)} \cite{llamaindex2025awd}
    \label{fig:agentic_doc_workflow}
\end{figure}

\textbf{Agentic Document Workflows (ADW)} \cite{llamaindex2025awd} extend traditional Retrieval-Augmented Generation (RAG) paradigms by enabling end-to-end knowledge work automation. These workflows orchestrate complex document-centric processes, integrating document parsing, retrieval, reasoning, and structured outputs with intelligent agents (see Figure \ref{fig:agentic_doc_workflow}). ADW systems address limitations of Intelligent Document Processing (IDP) and RAG by maintaining state, coordinating multi-step workflows, and applying domain-specific logic to documents.

\paragraph{Workflow}
\begin{enumerate}
    \item \textbf{Document Parsing and Information Structuring:}
    \begin{itemize}
        \item Documents are parsed using enterprise-grade tools (e.g., LlamaParse) to extract relevant data fields such as invoice numbers, dates, vendor information, line items, and payment terms.
        \item Structured data is organized for downstream processing.
    \end{itemize}
    
    \item \textbf{State Maintenance Across Processes:}
    \begin{itemize}
        \item The system maintains state about document context, ensuring consistency and relevance across multi-step workflows.
        \item Tracks the progression of the document through various processing stages.
    \end{itemize}
    
    \item \textbf{Knowledge Retrieval:}
    \begin{itemize}
        \item Relevant references are retrieved from external knowledge bases (e.g., LlamaCloud) or vector indexes.
        \item Retrieves real-time, domain-specific guidelines for enhanced decision-making.
    \end{itemize}
    
    \item \textbf{Agentic Orchestration:}
    \begin{itemize}
        \item Intelligent agents apply business rules, perform multi-hop reasoning, and generate actionable recommendations.
        \item Orchestrates components such as parsers, retrievers, and external APIs for seamless integration.
    \end{itemize}
    
    \item \textbf{Actionable Output Generation:}
    \begin{itemize}
        \item Outputs are presented in structured formats, tailored to specific use cases.
        \item Recommendations and extracted insights are synthesized into concise and actionable reports.
    \end{itemize}
\end{enumerate}

\begin{tcolorbox}[colframe=black, colback=white, boxrule=1pt, sharp corners=all, title=Use Case: Invoice Payments Workflow]
\textbf{Prompt:} Generate a payment recommendation report based on the submitted invoice and associated vendor contract terms.\\[0.2cm]

\textbf{System Process (ADW Workflow):}
\begin{enumerate}
    \item Parse the invoice to extract key details such as invoice number, date, vendor information, line items, and payment terms.
    \item Retrieve the corresponding vendor contract to verify payment terms and identify any applicable discounts or compliance requirements.
    \item Generate a payment recommendation report that includes original amount due, potential early payment discounts, budget impact analysis, and strategic payment actions.
\end{enumerate}

\textbf{Response:}
\textit{Integrated Response:}  
"Invoice INV-2025-045 for \$15,000.00 has been processed. An early payment discount of 2\% is available if paid by 2025-04-10, reducing the amount due to \$14,700.00. A bulk order discount of 5\% was applied as the subtotal exceeded \$10,000.00. It is recommended to approve early payment to save 2\% and ensure timely fund allocation for upcoming project phases."
\end{tcolorbox}

\paragraph{Key Features and Advantages}
\begin{itemize}
    \item \textbf{State Maintenance:} Tracks document context and workflow stage, ensuring consistency across processes.
    \item \textbf{Multi-Step Orchestration:} Handles complex workflows involving multiple components and external tools.
    \item \textbf{Domain-Specific Intelligence:} Applies tailored business rules and guidelines for precise recommendations.
    \item \textbf{Scalability:} Supports large-scale document processing with modular and dynamic agent integration.
    \item \textbf{Enhanced Productivity:} Automates repetitive tasks while augmenting human expertise in decision-making.
\end{itemize}

\section{Comparative Analysis of Agentic RAG Frameworks}

Table~\ref{tab:rag_comparison} provides a comprehensive comparative analysis of the three architectural frameworks: Traditional RAG, Agentic RAG, and Agentic Document Workflows (ADW). This analysis highlights their respective strengths, weaknesses, and best-fit scenarios, offering valuable insights into their applicability across diverse use cases.

\begin{table}[h]
\renewcommand{\arraystretch}{1}
    \centering
    \captionsetup{justification=centering, aboveskip=5pt, belowskip=10pt}
    \caption{Comparative Analysis: Traditional RAG vs Agentic RAG vs Agentic Document Workflows (ADW)}
    \begin{tabular}{|>{\centering\arraybackslash}p{3.4cm}|>{\centering\arraybackslash}p{3.4cm}|>{\centering\arraybackslash}p{3.4cm}|>{\centering\arraybackslash}p{4cm}|}
        \hline
        \rowcolor{lightgray} \textbf{Feature} & \textbf{Traditional RAG} & \textbf{Agentic RAG} & \textbf{Agentic Document Workflows (ADW)} \\
        \hline
        \textbf{Focus} & 
            Isolated retrieval and generation tasks &
            Multi-agent collaboration and reasoning &
            Document-centric end-to-end workflows \\
        \hline
        \textbf{Context Maintenance} & 
            Limited &
            Enabled through memory modules &
            Maintains state across multi-step workflows \\
        \hline
        \textbf{Dynamic Adaptability} & 
            Minimal &
            High &
            Tailored to document workflows \\
        \hline
        \textbf{Workflow Orchestration} & 
            Absent &
            Orchestrates multi-agent tasks &
            Integrates multi-step document processing \\
        \hline
        \textbf{Use of External Tools/APIs} & 
            Basic integration (e.g., retrieval tools) &
            Extends via tools like APIs and knowledge bases &
            Deeply integrates business rules and domain-specific tools \\
        \hline
        \textbf{Scalability} & 
            Limited to small datasets or queries &
            Scalable for multi-agent systems &
            Scales for multi-domain enterprise workflows \\
        \hline
        \textbf{Complex Reasoning} & 
            Basic (e.g., simple Q\&A) &
            Multi-step reasoning with agents &
            Structured reasoning across documents \\
        \hline
        \textbf{Primary Applications} & 
            QA systems, knowledge retrieval &
            Multi-domain knowledge and reasoning &
            Contract review, invoice processing, claims analysis \\
        \hline
        \textbf{Strengths} & 
            Simplicity, quick setup &
            High accuracy, collaborative reasoning &
            End-to-end automation, domain-specific intelligence \\
        \hline
        \textbf{Challenges} & 
            Poor contextual understanding &
            Coordination complexity &
            Resource overhead, domain standardization \\
        \hline
    \end{tabular}
    
    \label{tab:rag_comparison}
\end{table}

The comparative analysis underscores the evolutionary trajectory from Traditional RAG to Agentic RAG and further to Agentic Document Workflows (ADW). While Traditional RAG offers simplicity and ease of deployment for basic tasks, Agentic RAG introduces enhanced reasoning and scalability through multi-agent collaboration. ADW builds upon these advancements by providing robust, document-centric workflows that facilitate end-to-end automation and integration with domain-specific processes. Understanding the strengths and limitations of each framework is crucial for selecting the most appropriate architecture to meet specific application requirements and operational demands.

\section{Applications of Agentic RAG}

Agentic Retrieval-Augmented Generation (RAG) systems have demonstrated transformative potential across a variety of domains. By combining real-time data retrieval, generative capabilities, and autonomous decision-making, these systems address complex, dynamic, and multi-modal challenges. This section explores the key applications of Agentic RAG, providing detailed insights into how these systems are shaping industries such as customer support, healthcare, finance, education, legal workflows, and creative industries.

\subsection{Customer Support and Virtual Assistants}

Agentic RAG systems are revolutionizing customer support by enabling real-time, context-aware query resolution. Traditional chatbots and virtual assistants often rely on static knowledge bases, leading to generic or outdated responses. By contrast, Agentic RAG systems dynamically retrieve the most relevant information, adapt to the user's context, and generate personalized responses.

\textbf{Use Case: Twitch Ad Sales Enhancement} \cite{aws_twitch_agentic_rag}\\
For instance, Twitch leveraged an agentic workflow with RAG on Amazon Bedrock to streamline ad sales. The system dynamically retrieved advertiser data, historical campaign performance, and audience demographics to generate detailed ad proposals, significantly boosting operational efficiency.

\textbf{Key Benefits:}
\begin{itemize}
    \item \textbf{Improved Response Quality}: Personalized and context-aware replies enhance user engagement.
    \item \textbf{Operational Efficiency}: Reduces the workload on human support agents by automating complex queries.
    \item \textbf{Real-Time Adaptability}: Dynamically integrates evolving data, such as live service outages or pricing updates.
\end{itemize}

\subsection{Healthcare and Personalized Medicine}

In healthcare, the integration of patient-specific data with the latest medical research is critical for informed decision-making. Agentic RAG systems enable this by retrieving real-time clinical guidelines, medical literature, and patient history to assist clinicians in diagnostics and treatment planning.

\textbf{Use Case: Patient Case Summary} \cite{llamacloud_patient_case_summary}\\
Agentic RAG systems have been applied in generating patient case summaries. For example, by integrating electronic health records (EHR) and up-to-date medical literature, the system generates comprehensive summaries for clinicians to make faster and more informed decisions.

\textbf{Key Benefits:}
\begin{itemize}
    \item \textbf{Personalized Care}: Tailors recommendations to individual patient needs.
    \item \textbf{Time Efficiency}: Streamlines the retrieval of relevant research, saving valuable time for healthcare providers.
    \item \textbf{Accuracy}: Ensures recommendations are based on the latest evidence and patient-specific parameters.
\end{itemize}

\subsection{Legal and Contract Analysis}

Agentic RAG systems are redefining how legal workflows are conducted, offering tools for rapid document analysis and decision-making.

\textbf{Use Case: Contract Review} \cite{llamacloud_contract_review}\\
A legal agentic RAG system can analyze contracts, extract critical clauses, and identify potential risks. By combining semantic search capabilities with legal knowledge graphs, it automates the tedious process of contract review, ensuring compliance and mitigating risks.

\textbf{Key Benefits:}
\begin{itemize}
    \item \textbf{Risk Identification}: Automatically flags clauses that deviate from standard terms.
    \item \textbf{Efficiency}: Reduces the time spent on contract review processes.
    \item \textbf{Scalability}: Handles large volumes of contracts simultaneously.
\end{itemize}

\subsection{Finance and Risk Analysis}

Agentic RAG systems are transforming the finance industry by providing real-time insights for investment decisions, market analysis, and risk management. These systems integrate live data streams, historical trends, and predictive modeling to generate actionable outputs.

\textbf{Use Case: Auto Insurance Claims Processing} \cite{llamacloud_auto_insurance_claims}\\
In auto insurance, Agentic RAG can automate claim processing. For example, by retrieving policy details and combining them with accident data, it generates claim recommendations while ensuring compliance with regulatory requirements.

\textbf{Key Benefits:}
\begin{itemize}
    \item \textbf{Real-Time Analytics}: Delivers insights based on live market data.
    \item \textbf{Risk Mitigation}: Identifies potential risks using predictive analysis and multi-step reasoning.
    \item \textbf{Enhanced Decision-Making}: Combines historical and live data for comprehensive strategies.
\end{itemize}

\subsection{Education and Personalized Learning}

Education is another domain where Agentic RAG systems are making significant strides. These systems enable adaptive learning by generating explanations, study materials, and feedback tailored to the learner’s progress and preferences.

\textbf{Use Case: Research Paper Generation} \cite{llamacloud_research_paper_report}\\
In higher education, Agentic RAG has been used to assist researchers by synthesizing key findings from multiple sources. For instance, a researcher querying, “What are the latest advancements in quantum computing?” receives a concise summary enriched with references, enhancing the quality and efficiency of their work.

\textbf{Key Benefits:}
\begin{itemize}
    \item \textbf{Tailored Learning Paths}: Adapts content to individual student needs and performance levels.
    \item \textbf{Engaging Interactions}: Provides interactive explanations and personalized feedback.
    \item \textbf{Scalability}: Supports large-scale deployments for diverse educational environments.
\end{itemize}

\subsection{Graph-Enhanced Applications in Multimodal Workflows}

Graph-Enhanced Agentic RAG (GEAR) combines graph structures with retrieval mechanisms, making it particularly effective in multimodal workflows where interconnected data sources are essential.

\textbf{Use Case: Market Survey Generation}\\
GEAR enables the synthesis of text, images, and videos for marketing campaigns. For example, querying, “What are the emerging trends in eco-friendly products?” generates a detailed report enriched with customer preferences, competitor analysis, and multimedia content.

\textbf{Key Benefits:}
\begin{itemize}
    \item \textbf{Multi-Modal Capabilities}: Integrates text, image, and video data for comprehensive outputs.
    \item \textbf{Enhanced Creativity}: Generates innovative ideas and solutions for marketing and entertainment.
    \item \textbf{Dynamic Adaptability}: Adapts to evolving market trends and customer needs.
\end{itemize}

The applications of Agentic RAG systems span a wide range of industries, showcasing their versatility and transformative potential. From personalized customer support to adaptive education and graph-enhanced multimodal workflows, these systems address complex, dynamic, and knowledge-intensive challenges. By integrating retrieval, generation, and agentic intelligence, Agentic RAG systems are paving the way for next-generation AI applications.

\section{Tools and Frameworks for Agentic RAG}

Agentic Retrieval-Augmented Generation (RAG) systems represent a significant evolution in combining retrieval, generation, and agentic intelligence. These systems extend the capabilities of traditional RAG by integrating decision-making, query reformulation, and adaptive workflows. The following tools and frameworks provide robust support for developing Agentic RAG systems, addressing the complex requirements of real-world applications.

\textbf{Key Tools and Frameworks:}
\begin{itemize}
    \item \textbf{LangChain and LangGraph:} LangChain \cite{langgraph_agentic_rag} provides modular components for building RAG pipelines, seamlessly integrating retrievers, generators, and external tools. LangGraph complements this by introducing graph-based workflows that support loops, state persistence, and human-in-the-loop interactions, enabling sophisticated orchestration and self-correction mechanisms in agentic systems.
    \item \textbf{LlamaIndex:} LlamaIndex's \cite{agentic_rag_llamaindex} Agentic Document Workflows (ADW) enable end-to-end automation of document processing, retrieval, and structured reasoning. It introduces a meta-agent architecture where sub-agents manage smaller document sets, coordinating through a top-level agent for tasks such as compliance analysis and contextual understanding.
    \item \textbf{Hugging Face Transformers and Qdrant:} Hugging Face \cite{huggingface_agent_rag} offers pre-trained models for embedding and generation tasks, while Qdrant \cite{qdrant_agentic_rag} enhances retrieval workflows with adaptive vector search capabilities, allowing agents to optimize performance by dynamically switching between sparse and dense vector methods.
    \item \textbf{CrewAI and AutoGen:} These frameworks emphasize multi-agent architectures. CrewAI \cite{crewAI} supports hierarchical and sequential processes, robust memory systems, and tool integrations. AG2 \cite{ag2} (formerly knows as AutoGen  \cite{wu2023autogen, zhang2024training}) excels in multi-agent collaboration with advanced support for code generation, tool execution, and decision-making.
    \item \textbf{OpenAI Swarm Framework:} An educational framework designed for ergonomic, lightweight multi-agent orchestration \cite{openai_swarm}, emphasizing agent autonomy and structured collaboration.
    \item \textbf{Agentic RAG with Vertex AI:} Developed by Google, Vertex AI  \cite{llamaindex_vertex_ai_agentic_rag} integrates seamlessly with Agentic Retrieval-Augmented Generation (RAG), providing a platform to build, deploy, and scale machine learning models while leveraging advanced AI capabilities for robust, contextually aware retrieval and decision-making workflows.
    \item \textbf{Semantic Kernel:}
    Semantic Kernel \cite{ microsoft2025semantickernel, microsoft2025semantickernelgithub} is an open-source SDK by Microsoft that integrates large language models (LLMs) into applications. It supports agentic patterns, enabling the creation of autonomous AI agents for natural language understanding, task automation, and decision-making. It has been used in scenarios like ServiceNow’s P1 incident management to facilitate real-time collaboration, automate task execution, and retrieve contextual information seamlessly
   \item \textbf{Amazon Bedrock for Agentic RAG:} Amazon Bedrock \cite{aws_twitch_agentic_rag} provides a robust platform for implementing Agentic Retrieval-Augmented Generation (RAG) workflows.
 \item \textbf{IBM Watson and Agentic RAG:} IBM's watsonx.ai \cite{ibm_granite_agentic_rag} supports building Agentic RAG systems, exemplified by using the Granite-3-8B-Instruct model to answer complex queries by integrating external information and enhancing response accuracy.

    \item \textbf{Neo4j and Vector Databases:} Neo4j, a prominent open-source graph database, excels in handling complex relationships and semantic queries. Alongside Neo4j, vector databases like Weaviate, Pinecone, Milvus, and Qdrant provide efficient similarity search and retrieval capabilities, forming the backbone of high-performance Agentic Retrieval-Augmented Generation (RAG) workflows.

\end{itemize}

\section{Comparative Analysis of Agentic RAG Frameworks}

Table~\ref{tab:rag_comparison} compares Traditional RAG and Agentic RAG across key dimensions. Traditional RAG offers simplicity for basic tasks, while Agentic RAG introduces enhanced reasoning and scalability through multi-agent collaboration. Understanding their respective strengths and limitations is essential for selecting an architecture that meets specific application requirements.

\subsection{Comparative Insights and Design Trade-offs}
A taxonomy-driven analysis reveals key trade-offs beyond feature-level distinctions. Single-agent architectures favor simplicity and low latency but struggle with multi-domain reasoning. Multi-agent systems improve scalability through parallelism at the cost of coordination overhead. Hierarchical and corrective architectures enhance reliability through strategic oversight but incur higher latency. Graph-based systems enable structured relational reasoning while introducing dependencies on knowledge quality. Table~\ref{tab:taxonomy_comparison} maps these architectures to taxonomy dimensions and design trade-offs.


\begin{table}[h]
\renewcommand{\arraystretch}{1.3}
\centering
\captionsetup{justification=centering, aboveskip=5pt, belowskip=10pt}
\caption{Taxonomy-Driven Comparison of Agentic RAG Architectures}
\label{tab:taxonomy_comparison}

\begin{tabular}{|
>{\centering\arraybackslash}p{3cm}|
>{\centering\arraybackslash}p{3cm}|
>{\centering\arraybackslash}p{3cm}|
>{\centering\arraybackslash}p{3.2cm}|
>{\centering\arraybackslash}p{3.2cm}|}
\hline

\rowcolor{lightgray}
\textbf{Dimension} &
\textbf{Single-Agent RAG} &
\textbf{Multi-Agent RAG} &
\textbf{Hierarchical / Corrective RAG} &
\textbf{Graph-Based / Document-Centric RAG} \\
\hline

\textbf{Agent Cardinality} &
Single &
Multiple &
Multiple (tiered) &
Multiple \\
\hline

\textbf{Control Structure} &
Centralized &
Flat &
Hierarchical / evaluator-based &
Adaptive / structured \\
\hline

\textbf{Knowledge Representation} &
Unstructured &
Unstructured or hybrid &
Unstructured or hybrid &
Graph-based or document-centric \\
\hline

\textbf{Knowledge Dependency} &
Low &
Moderate &
Moderate &
High, sensitive to graph quality \\
\hline

\textbf{Reasoning Mode} &
Implicit, LLM-driven &
Distributed across agents &
Strategic with supervisory control &
Relational or symbolic + LLM \\
\hline

\textbf{Reasoning Depth} &
Low to moderate &
Moderate to high &
High &
High, multi-hop reasoning \\
\hline

\textbf{Retrieval Adaptivity} &
Limited &
High &
High &
High \\
\hline

\textbf{Scalability} &
Moderate &
High &
Moderate &
Moderate \\
\hline

\textbf{Latency} &
Low &
Moderate &
High &
Moderate to high \\
\hline

\textbf{Engineering Complexity} &
Low &
Moderate &
High &
High \\
\hline

\textbf{Best-Fit Scenarios} &
Simple QA and routing &
Multi-domain synthesis &
High-stakes, reliability-critical tasks &
Knowledge-intensive, relational domains \\
\hline

\end{tabular}
\end{table}

\section{Lessons Learned and Practical Guidance}

Based on the analysis of paradigms, taxonomies, and real-world use cases, this section distills practical insights for designing, evaluating, and deploying Agentic RAG systems.


\subsection{Agentic RAG Is Not Always the Right Default}
Agentic RAG should not be viewed as a universal replacement for traditional RAG. While it offers superior adaptability and multi-step reasoning, it also introduces coordination complexity, latency, and computational cost. For simple fact retrieval or well-scoped queries, modular RAG pipelines provide sufficient performance with lower overhead. Practitioners should adopt agentic designs selectively, guided by task complexity.

\subsection{Architectural Choice Strongly Shapes System Behavior}
Different architectures lead to fundamentally different behaviors. Single-agent systems favor simplicity; multi-agent systems enable parallelism but require orchestration; hierarchical architectures introduce oversight at the expense of complexity. Architectural decisions implicitly encode assumptions about control, trust, and error tolerance---corrective workflows improve accuracy but add latency, while adaptive architectures trade correctness guarantees for responsiveness.


\subsection{Retrieval Quality Remains the Primary Bottleneck}
Agentic reasoning cannot compensate for consistently poor retrieval. Failures often originate from inadequate retrieval coverage, poorly constructed indexes, or insufficient integration of structured and unstructured knowledge. Investing in robust retrieval pipelines and high-quality indexing before introducing agentic complexity remains essential.


\subsection{Agent Autonomy Requires Explicit Constraints}
Unrestricted autonomy risks excessive tool invocation, redundant reasoning loops, or misaligned objectives. Effective systems balance autonomy with bounded planning horizons, predefined tool access policies, and explicit stopping criteria. Constrained autonomy yields more reliable outcomes, especially in production environments.

\subsection{Evaluation Must Account for Process, Not Just Outcomes}
Existing benchmarks focus on output quality with limited visibility into intermediate decisions. Meaningful evaluation requires process-level metrics capturing reasoning efficiency, tool usage patterns, and adaptation to changing contexts. Without such evaluation, improvements risk being anecdotal rather than systematic.



\subsection{Domain Knowledge Significantly Amplifies Agentic Benefits}
Agentic RAG achieves its strongest gains in domains with structured knowledge and explicit constraints. Healthcare, finance, and legal analysis particularly benefit from combining retrieval with rule-based reasoning and graph-structured knowledge. Open-domain tasks show more modest gains, underscoring domain modeling as a complementary component of Agentic RAG design.


\subsection{Toward Responsible Deployment of Agentic RAG}
Agentic RAG introduces challenges in transparency, accountability, and trust. Multi-agent collaboration complicates error attribution, while autonomous tool use raises safety concerns. Responsible deployment requires governance mechanisms, human oversight, and clear operational boundaries. Explainability, traceability, and auditability must be first-class design goals, particularly for high-stakes applications.


\section{Benchmarks and Datasets}
Current benchmarks~\cite{Ferrag2025FromLLMToAutonomousAI} and datasets evaluate RAG systems across retrieval, reasoning, and generation. Table~\ref{tab:rag_tasks_datasets_with_references} summarizes key datasets by downstream task. The following benchmarks are particularly relevant:

\begin{itemize}
    \item \textit{BEIR (Benchmarking Information Retrieval):} A benchmark for evaluating embedding models across 17 datasets in bioinformatics, finance, and question answering  \cite{thakur2021beirheterogenousbenchmarkzeroshot}.

    \item \textit{MS MARCO (Microsoft Machine Reading Comprehension):} 
    Passage ranking and question answering, widely used for dense retrieval evaluation in RAG systems~\cite{bajaj2018msmarco}.

    \item \textit{TREC (Text REtrieval Conference, Deep Learning Track):} Provides datasets for passage and document retrieval, emphasizing ranking quality in retrieval pipelines \cite{craswell2023overview}.

    \item \textit{MuSiQue (Multihop Sequential Questioning):} A benchmark for multi-document reasoning, emphasizing retrieval and synthesis across contexts~\cite{trivedi2022musiquemultihopquestionssinglehop}.

    \item \textit{2WikiMultihopQA:} Designed for multihop QA over Wikipedia articles with cross-source integration\cite{ho2020constructingmultihopqadataset}.

    \item \textit{AgentG} (Agentic RAG for Knowledge Fusion)~\cite{lee2024agentg}: Tailored for agenticRAG tasks, this benchmark assesses dynamic information synthesis across knowledge bases.

    Tailored for Agentic RAG, assessing dynamic information synthesis across knowledge bases.

    \item \textit{HotpotQA:} A multihop QA benchmark requiring retrieval and reasoning over interconnected contexts, suitable for evaluating complex RAG workflows \cite{yang2018hotpotqa}.

    \item \textit{RAGBench:} A large-scale, explainable benchmark with 100,000 examples across domains, using the TRACe framework for actionable RAG evaluation \cite{friel2024ragbenchexplainablebenchmarkretrievalaugmented}.
    
    \item \textit{BERGEN (Benchmarking Retrieval-Augmented Generation):} A library for systematically benchmarking RAG systems through standardized experiments \cite{rau2024bergenbenchmarkinglibraryretrievalaugmented}.
    
    \item \textit{FlashRAG Toolkit:} Implements multiple RAG methods and benchmark datasets to support efficient and standardized evaluation~\cite{jin2024flashragmodulartoolkitefficient}.
    
    \item \textit{GNN-RAG:} Evaluates graph-based RAG systems on node- and edge-level prediction tasks, focusing on retrieval quality and reasoning performance in knowledge graph question answering \cite{mavromatis2024gnnraggraphneuralretrieval}.
\end{itemize}

\begin{table*}[!t]
\caption{Downstream Tasks and Datasets for RAG Evaluation (Adapted from \cite{gao2024retrievalaugmentedgenerationlargelanguage})}
\label{tab:rag_tasks_datasets_with_references}
\centering
\renewcommand{\arraystretch}{1}
\begin{tabular}{p{2.3cm} p{3.3cm} p{10.6cm}}
\hline
\multicolumn{1}{c}{\textbf{Category}} &
\multicolumn{1}{c}{\textbf{Task Type}} &
\multicolumn{1}{c}{\textbf{Datasets and References}} \\
\hline

\multirow{5}{*}{\textbf{QA}} 
& Single-hop QA 
& Natural Questions (NQ) \cite{kwiatkowski2019naturalquestions}, TriviaQA \cite{joshi2017triviaqa}, SQuAD \cite{rajpurkar2016squad}, Web Questions (WebQ) \cite{berant2013webquestions}, PopQA \cite{mallen-etal-2023-trust}, MS MARCO \cite{bajaj2018msmarco} \\

& Multi-hop QA 
& HotpotQA \cite{yang2018hotpotqa}, 2WikiMultiHopQA \cite{ho2020constructingmultihopqadataset}, MuSiQue \cite{trivedi2022musiquemultihopquestionssinglehop} \\

& Long-form QA 
& ELI5 \cite{fan2019eli5}, NarrativeQA (NQA) \cite{kovcisky2018narrativeqa}, ASQA \cite{stelmakh2023asqafactoidquestionsmeet}, QMSum \cite{zhong2021qmsum} \\

& Domain-specific QA 
& Qasper \cite{dasigi2021qasper}, COVID-QA \cite{moller2020covidqa}, CMB/MMCU Medical \cite{wang2024cmbcomprehensivemedicalbenchmark} \\

& Multi-choice QA 
& QuALITY \cite{pang2022qualityquestionansweringlong}, ARC, CommonsenseQA \cite{talmor-etal-2019-commonsenseqa} \\
\hline

\multirow{2}{*}{\textbf{Graph-based QA}} 
& Graph QA 
& GraphQA \cite{he2024gretrieverretrievalaugmentedgenerationtextual} \\

& Event Argument Extraction 
& WikiEvent \cite{li2021documentleveleventargumentextraction}, RAMS \cite{ebner2020multisentenceargumentlinking} \\
\hline

\multirow{3}{*}{\textbf{Dialog}} 
& Open-domain Dialog 
& Wizard of Wikipedia (WoW) \cite{dinan2019wizardwikipediaknowledgepoweredconversational} \\

& Personalized Dialog 
& KBP \cite{wang2023largelanguagemodelssource}, DuleMon \cite{xu2022longtimeseeopendomain} \\

& Task-oriented Dialog 
& CamRest \cite{wen-etal-2016-conditional} \\
\hline

\textbf{Recommendation} 
& Personalized Content 
& Amazon Datasets (Toys, Sports, Beauty) \cite{runing2016} \\
\hline

\multirow{3}{*}{\textbf{Reasoning}} 
& Commonsense Reasoning 
& HellaSwag \cite{zellers-etal-2019-hellaswag}, CommonsenseQA \cite{talmor-etal-2019-commonsenseqa} \\

& Chain-of-Thought Reasoning 
& CoT Collection \cite{kim2023cotcollectionimprovingzeroshot} \\

& Complex Reasoning 
& CSQA \cite{csqa} \\
\hline

\multirow{3}{*}{\textbf{Others}} 
& Language Understanding 
& MMLU, WikiText-103 \cite{kwiatkowski2019naturalquestions} \\

& Fact Checking and Verification 
& FEVER \cite{thorne-etal-2018-fever}, PubHealth \cite{kotonya2020explainableautomatedfactcheckingpublic} \\

& Strategy QA 
& StrategyQA \cite{geva2021didaristotleuselaptop} \\
\hline

\multirow{2}{*}{\textbf{Summarization}} 
& Text Summarization 
& WikiASP \cite{hayashi2020wikiaspdatasetmultidomainaspectbased}, XSum \cite{narayan2018dontdetailsjustsummary} \\

& Long-form Summarization 
& NarrativeQA (NQA) \cite{kovcisky2018narrativeqa}, QMSum \cite{zhong2021qmsum} \\
\hline

\textbf{Text Generation} 
& Biography 
& Biography Dataset \\ 
\hline

\multirow{2}{*}{\textbf{Text Classification}} 
& Sentiment Analysis 
& SST-2 \cite{socher-etal-2013-recursive} \\

& General Classification 
& VioLens \cite{saha-etal-2023-vio}, TREC \cite{craswell2023overview} \\
\hline

\textbf{Code Search} 
& Programming Search 
& CodeSearchNet \cite{husain2020codesearchnetchallengeevaluatingstate} \\
\hline

\multirow{2}{*}{\textbf{Robustness}} 
& Retrieval Robustness 
& NoMIRACL \cite{thakur2024knowingdontknowmultilingual} \\

& Language Modeling Robustness 
& WikiText-103 \cite{merity2016pointersentinelmixturemodels} \\
\hline

\textbf{Math} 
& Math Reasoning 
& GSM8K \cite{cobbe2021gsm8k} \\
\hline

\textbf{Machine Translation} 
& Translation Tasks 
& JRC-Acquis \cite{steinberger-etal-2006-jrc} \\
\hline
\end{tabular}
\end{table*}

\section{Open Research Issues and Future Challenges in Agentic RAG}
Despite rapid progress in Agentic RAG, the field remains at an early stage of maturity. While existing systems demonstrate clear advantages over traditional RAG pipelines, they also expose a range of unresolved research challenges that must be addressed to enable robust, scalable, and trustworthy deployment in real-world settings. This section outlines key open research issues and articulates a long-term research vision for Agentic RAG systems.

\subsection{Agent Coordination, Control, and Emergent Behavior}
As systems scale to multi-agent and hierarchical frameworks, emergent behaviors become difficult to predict. Current orchestration relies on heuristic rules and prompt engineering with limited convergence guarantees. Future research must address coordination under partial observability and conflicting objectives, formally model inter-agent dependencies, prevent cascading failures, and ensure verifiable collaborative outcomes.


\subsection{Evaluation Methodologies Beyond Output Quality}
Output-level metrics (answer correctness, retrieval accuracy) are insufficient for Agentic RAG, where intermediate decisions shape performance. Standardized benchmarks evaluating reasoning trajectories, planning depth, adaptability, robustness under noisy retrieval, and cost efficiency are needed. Process-aware evaluation assessing \textit{how} answers are produced is essential for systematic progress.

\subsection{Memory Management and Long-Term Adaptation}
Long-term memory design remains open challenge. Persistent memory risks knowledge drift and bias reinforcement, while frequent updates can amplify hallucinations. Key questions include balancing persistence with adaptability, selective retention, and reconciling external knowledge with stored agent experiences.

\subsection{Computational Cost, Efficiency, and Sustainability}
Multi-agent collaboration and iterative retrieval increase latency and resource consumption. Future research must explore cost-aware planning, adaptive inference, and lightweight coordination---dynamically adjusting active agents, selectively invoking tools, and optimizing retrieval depth based on task complexity.

\subsection{Safety, Trust, and Governance in Autonomous RAG Systems}
Autonomous agents may introduce failure modes including unauthorized actions and biased decisions. In multi-agent settings, tracing responsibility is particularly challenging. Open directions include transparent decision-tracing, human-in-the-loop oversight, and governance frameworks ensuring alignment between agent objectives and human intent in high-stakes applications.

\subsection{Generalization Across Domains and Tasks}
Most existing Agentic RAG systems are evaluated in domain-specific settings or carefully curated benchmarks. It remains unclear how well these systems generalize across domains with differing data characteristics, regulatory constraints, or reasoning requirements. Designing agents that can adapt their retrieval and reasoning strategies without extensive task-specific tuning is an open problem. Future work should focus on developing domain-agnostic agentic abstractions, transferable planning strategies, and adaptive retrieval policies that generalize beyond narrow application contexts. Achieving robust cross-domain generalization will be essential for realizing the full potential of Agentic RAG as a general-purpose reasoning framework.

\section{Conclusion}
Agentic RAG represents a significant evolution beyond static RAG pipelines toward autonomous, adaptive reasoning frameworks. This survey examined RAG foundations, agentic intelligence integration, and the architectural patterns defining modern Agentic RAG systems. By organizing approaches through a principled taxonomy based on agent cardinality, control structure, autonomy, and knowledge representation, this work provides a structured view of a fragmented landscape. The analysis demonstrates that no single architecture is universally optimal; effectiveness depends on aligning complexity with task requirements. Agentic RAG offers clear advantages for complex, knowledge-intensive tasks in healthcare, finance, education, and enterprise processing, but these benefits require high-quality retrieval foundations, disciplined autonomy, and careful orchestration.

Substantial open challenges remain in multi-agent coordination, process-aware evaluation, memory management, computational efficiency, and governance. This survey consolidates existing work, introduces a unifying framework, and articulates design trade-offs and research directions to advance the field of Agentic RAG.

\begin{table*}[ht]
\renewcommand{\arraystretch}{1.3}
\centering
\caption{Downstream Tasks and Datasets for RAG Evaluation (Adapted from \cite{gao2024retrievalaugmentedgenerationlargelanguage}}
\label{tab:rag_tasks_datasets_with_references}
\begin{tabular}{|l|l|p{8cm}|}
\hline
\rowcolor{lightgray} \textbf{Category}              & \textbf{Task Type}            & \textbf{Datasets and References}                                                                  \\ \hline
\multirow{5}{*}{\textbf{QA}}   & Single-hop QA                 & Natural Questions (NQ) \cite{kwiatkowski2019naturalquestions}, TriviaQA \cite{joshi2017triviaqa}, SQuAD \cite{rajpurkar2016squad}, Web Questions (WebQ) \cite{berant2013webquestions}, PopQA \cite{mallen-etal-2023-trust}, MS MARCO \cite{bajaj2018msmarco}         \\ \cline{2-3}
                               & Multi-hop QA                  & HotpotQA \cite{yang2018hotpotqa}, 2WikiMultiHopQA \cite{ho2020constructingmultihopqadataset}, MuSiQue \cite{trivedi2022musiquemultihopquestionssinglehop}                                                     \\ \cline{2-3}
                               & Long-form QA                 & ELI5 \cite{fan2019eli5}, NarrativeQA (NQA) \cite{kovcisky2018narrativeqa}, ASQA \cite{stelmakh2023asqafactoidquestionsmeet}, QMSum \cite{zhong2021qmsum}                                                   \\ \cline{2-3}
                               & Domain-specific QA            & Qasper \cite{dasigi2021qasper}, COVID-QA \cite{moller2020covidqa}, CMB/MMCU Medical \cite{wang2024cmbcomprehensivemedicalbenchmark}                                              \\ \cline{2-3}
                               & Multi-choice QA              & QuALITY \cite{pang2022qualityquestionansweringlong}, ARC (No reference available), CommonsenseQA \cite{talmor-etal-2019-commonsenseqa}                                                         \\ \hline
\multirow{2}{*}{\textbf{Graph-based QA}} 
                               & Graph QA                      & GraphQA \cite{he2024gretrieverretrievalaugmentedgenerationtextual}                                                                             \\ \cline{2-3}
                               & Event Argument Extraction     & WikiEvent \cite{li2021documentleveleventargumentextraction}, RAMS \cite{ebner2020multisentenceargumentlinking}                                                                     \\ \hline
\multirow{3}{*}{\textbf{Dialog}} & Open-domain Dialog           & Wizard of Wikipedia (WoW) \cite{dinan2019wizardwikipediaknowledgepoweredconversational}                                                             \\ \cline{2-3}
                               & Personalized Dialog          & KBP \cite{wang2023largelanguagemodelssource}, DuleMon \cite{xu2022longtimeseeopendomain}                                                                          \\ \cline{2-3}
                               & Task-oriented Dialog         & CamRest \cite{wen-etal-2016-conditional}                                                                              \\ \hline
\textbf{Recommendation} & Personalized Content      & Amazon Datasets (Toys, Sports, Beauty) \cite{runing2016}                  \\ \hline
\multirow{3}{*}{\textbf{Reasoning}} & Commonsense Reasoning      & HellaSwag \cite{ zellers-etal-2019-hellaswag}, CommonsenseQA \cite{talmor-etal-2019-commonsenseqa}                        \\ \cline{2-3}
                               & CoT Reasoning                & CoT Reasoning \cite{kim2023cotcollectionimprovingzeroshot}                                                                    \\ \cline{2-3}
                               & Complex Reasoning            & CSQA \cite{csqa}                                                                                 \\ \hline
\multirow{3}{*}{\textbf{Others}} & Language Understanding       & MMLU (No reference available), WikiText-103 \cite{kwiatkowski2019naturalquestions}                                                                   \\ \cline{2-3}
                               & Fact Checking/Verification    & FEVER \cite{thorne-etal-2018-fever}, PubHealth \cite{kotonya2020explainableautomatedfactcheckingpublic}                                                                     \\ \cline{2-3}
                               & Strategy QA                  & StrategyQA \cite{geva2021didaristotleuselaptop}                                                                           \\ \hline
\multirow{2}{*}{\textbf{Summarization}} & Text Summarization        & WikiASP \cite{hayashi2020wikiaspdatasetmultidomainaspectbased}, XSum \cite{narayan2018dontdetailsjustsummary}                                                                        \\ \cline{2-3}
                               & Long-form Summarization      & NarrativeQA (NQA) \cite{kovcisky2018narrativeqa}, QMSum \cite{zhong2021qmsum}                                                             \\ \hline
\textbf{Text Generation}       & Biography                     & Biography Dataset (No reference available)                                                                     \\ \hline
\multirow{2}{*}{\textbf{Text Classification}} 
                               & Sentiment Analysis           & SST-2 \cite{socher-etal-2013-recursive}                                                                                \\ \cline{2-3}
                               & General Classification       & VioLens\cite{saha-etal-2023-vio}, TREC \cite{craswell2023overview}                                                                        \\ \hline
\textbf{Code Search}           & Programming Search            & CodeSearchNet \cite{husain2020codesearchnetchallengeevaluatingstate}                                                                     \\ \hline
\multirow{2}{*}{\textbf{Robustness}} & Retrieval Robustness       & NoMIRACL \cite{thakur2024knowingdontknowmultilingual}                                                                        \\ \cline{2-3}
                               & Language Modeling Robustness & WikiText-103 \cite{merity2016pointersentinelmixturemodels}                                                                         \\ \hline
\textbf{Math}                  & Math Reasoning                & GSM8K \cite{cobbe2021gsm8k}                                                                               \\ \hline
\textbf{Machine Translation}   & Translation Tasks             & JRC-Acquis \cite{steinberger-etal-2006-jrc}                                                                          \\ \hline
\end{tabular}
\end{table*}

\bibliographystyle{unsrt}  
\bibliography{references}

\end{document}